\documentclass[11pt]{article}

\usepackage[preprint]{acl}

\usepackage{times}
\usepackage{latexsym}

\usepackage[T1]{fontenc}
\usepackage[utf8]{inputenc}

\usepackage{microtype}
\usepackage{subfig}

\usepackage{inconsolata}

\usepackage{graphicx}
\usepackage{booktabs}
\usepackage{multirow}
\usepackage{amsmath}
\usepackage{caption}
\usepackage{subcaption}
\usepackage{subfig}
\usepackage{graphicx}
\usepackage{enumitem}

\usepackage[table]{xcolor}
\usepackage{array}
\usepackage{amssymb}

\definecolor{myorange}{RGB}{230,140,40}
\definecolor{myviolet}{RGB}{148,87,179}
\definecolor{mygreen}{RGB}{40,160,80}
\definecolor{myred}{RGB}{200,50,50}
\definecolor{myyellow}{RGB}{200,160,30}
\definecolor{myblue}{RGB}{40,100,200}

\newcommand{\stars}[1]{\textcolor{#1}{\ensuremath{\bigstar}}}
\newcommand{\xmark}{\textcolor{red}{\ensuremath{\times}}}

\title{Benchmarking Speech-to-Speech Translation Models}

\author{
    Alkis Koudounas\textsuperscript{$\dagger$} \quad
    Hayato Futami\textsuperscript{$\dagger$} \quad
    Quentin Jodelet\textsuperscript{$\dagger$} \\
    \textbf{Osamu Take\textsuperscript{$\dagger$} \quad
    Shinji Watanabe\textsuperscript{$\ddagger$} \quad   
    Emiru Tsunoo\textsuperscript{$\dagger$}} \\
  \textsuperscript{$\dagger$}Sony Group Corporation, Japan \quad
  \textsuperscript{$\ddagger$}Carnegie Mellon University, USA \\
  \texttt{Alkis.Koudounas@sony.com}}

\begin{document}

\maketitle

\begin{abstract}
Speech-to-speech translation (S2ST) has advanced rapidly, but offline evaluation lacks a unified protocol: studies report non-overlapping metric subsets, preventing direct comparisons.
We introduce COMPASS, a unified and reproducible benchmarking framework integrating 46 metrics across eight dimensions, and deploy it on 1,248 model-language configurations from FLEURS and CVSS, spanning cascaded and end-to-end architectures over ten language pairs.
Architectures exhibit complementary strengths: best-vs-worst gaps exceed 30\% on naturalness and speaker preservation but remain within a few points on translation quality, so single-metric rankings systematically misrepresent system quality.
Correlation filtering reduces 46 metrics to 10 per direction, with three axes requiring different metrics across X$\to$EN and EN$\to$X (e.g., TER/UTMOS vs.\ ChrF++/NISQA-MOS); these subsets preserve rankings (Spearman's $\rho>0.80$) while cutting evaluation time by $\approx 2.5\times$.
Human validation across dubbing, podcasts, and medical domains shows standalone MOS predictors fail to predict listener preference, while top domain-specific metrics correlate with human judgment ($\rho \geq 0.90$).
We release COMPASS as a foundation for domain-aware S2ST evaluation.
\end{abstract}

\begin{table*}[!t]
    \centering
    \renewcommand{\arraystretch}{0.83}
    \setlength{\tabcolsep}{6pt}
    \resizebox{2\columnwidth}{!}{%
    \begin{tabular}{l c c c c c c}
    \toprule
    \multirow{2}{*}{\textbf{Metric dimension}} 
        & \textbf{Live} & \textbf{Video}   & \textbf{Lecture/} & \textbf{Conversation/} & \textbf{Podcast*/}   & \textbf{News/} \\[-1pt]
        & \textbf{interpreting}    & \textbf{dubbing*} & \textbf{EDU}      & \textbf{Medical*}       & \textbf{Audiobook}  & \textbf{Broadcast} \\
    \midrule
    \textcolor{myblue}{\textbf{Translation Quality}} 
      & \stars{myblue}\stars{myblue}\stars{myblue} 
      & \stars{myblue}\stars{myblue} 
      & \stars{myblue}\stars{myblue}\stars{myblue} 
      & \stars{myblue}\stars{myblue}\stars{myblue} 
      & \stars{myblue}\stars{myblue} 
      & \stars{myblue}\stars{myblue}\stars{myblue} \\
    
    \textcolor{mygreen}{\textbf{Naturalness}} 
      & \stars{mygreen}\stars{mygreen} 
      & \stars{mygreen}\stars{mygreen}\stars{mygreen} 
      & \stars{mygreen}\stars{mygreen} 
      & \stars{mygreen}\stars{mygreen}\stars{mygreen} 
      & \stars{mygreen}\stars{mygreen}\stars{mygreen} 
      & \stars{mygreen}\stars{mygreen} \\
    
    \textcolor{myred}{\textbf{Speaker consistency}} 
      & \stars{myred}\stars{myred}
      & \stars{myred}\stars{myred}\stars{myred} 
      & \stars{myred}\stars{myred} 
      & \stars{myred}\stars{myred} 
      & \stars{myred}\stars{myred}\stars{myred} 
      & \stars{myred}\stars{myred} \\
    
    \textcolor{myyellow}{\textbf{Prosody / Emotion preservation}} 
      & \stars{myyellow}\stars{myyellow} 
      & \stars{myyellow}\stars{myyellow}\stars{myyellow} 
      & \stars{myyellow}\stars{myyellow}\stars{myyellow} 
      & \stars{myyellow}\stars{myyellow} 
      & \stars{myyellow}\stars{myyellow}\stars{myyellow} 
      & \stars{myyellow}\stars{myyellow} \\
    
    \textcolor{myorange}{\textbf{Isochrony / Isometry}} 
      & \stars{myorange}\stars{myorange} 
      & \stars{myorange}\stars{myorange}\stars{myorange} 
      & \xmark 
      & \xmark 
      & \stars{myorange}\stars{myorange} 
      & \stars{myorange} \\
    
    \textcolor{myviolet}{\textbf{Lip-syncing}} 
      & \xmark 
      & \stars{myviolet}\stars{myviolet}\stars{myviolet} 
      & \xmark 
      & \xmark 
      & \xmark 
      & \xmark \\
    \bottomrule
    \end{tabular}}
    \caption{Taxonomy relevance across S2ST use cases (${\star\star\star}$: critical, ${\star}$: secondary, ${\times}$: irrelevant). Levels reflect structural application constraints (e.g., visual sync for dubbing vs. strict semantic fidelity for medical dialogue). We selected three domains with * for human evaluation to see correlation with COMPASS.}
    \label{tab:domain_metrics}
    \vspace{-3mm}
\end{table*}

\section{Introduction}
Speech-to-speech translation (S2ST) is a core technology for communication \cite{lee2022direct, ahmad-etal-2024-findings, agostinelli-etal-2025-findings}, powering applications from video dubbing and live broadcasting to real-time conversation and accessibility.
Recent progress spans both cascaded pipelines chaining automatic speech recognition (ASR), machine translation (MT), and text-to-speech (TTS) synthesis \cite{bentivogli2021cascade}, and end-to-end models \cite{barrault2023seamlessm4t, xu2025qwen25omnitechnicalreport, xu2025qwen3}.

However, while streaming and simultaneous setups have advanced~\cite{gaido2025simulstream}, offline S2ST evaluation still lacks a widely adopted unified protocol.
A robust system must ensure accurate translation, natural speech, speaker and emotion preservation, and temporal alignment. Since each dimension uses its own metric family, studies often report non-overlapping subsets, preventing cross-paper comparisons.
No standardized toolkit ensures consistent metric computation, and it remains unclear which metrics matter per use case.
As Table~\ref{tab:domain_metrics} shows, while translation accuracy matters across the board, deployment scenarios differ in their additional priorities: dubbing requires lip-syncing and isochrony, podcasts emphasize speaker consistency and naturalness, and medical settings demand the strictest accuracy.
No uniform metric set fits all use cases, yet existing frameworks fail to operationalize these domain distinctions, leaving practitioners unable to choose the right system.

To close this gap, we introduce the first \textbf{COMP}rehensive \textbf{AS}sessment \textbf{S}uite (COMPASS) for benchmarking offline S2ST systems.
COMPASS unifies 40+ metrics across 8 dimensions: translation quality, evaluated both on the intermediate translated text (to isolate translation from synthesis errors) and on ASR transcripts of the synthesized speech (to capture end-to-end output quality), audio naturalness, speaker consistency, prosody \& emotion, isochrony, isometry, and lip sync.
The framework is modular, letting users select metrics per use case, and reproducible via a single toolkit.
We deploy COMPASS in the largest empirical S2ST evaluation study to date, benchmarking cascaded and end-to-end architectures across 1,248 model-language configurations, ten languages, and both directions (EN$\leftrightarrow$X) on FLEURS~\cite{conneau2023fleurs} and CVSS~\cite{jia2022cvss}.
Our analysis shows no architecture dominates: best-vs-worst gaps exceed 30\% on naturalness and speaker preservation yet stay within a few points on translation quality, making single-metric rankings misleading.
Many widely used metrics are also highly correlated; through correlation-based filtering, we identify compact subsets preserving discriminative power, with X$\to$EN and EN$\to$X requiring different metrics on three axes (text quality: TER vs.\ ChrF++; naturalness: UTMOS vs.\ NISQA-MOS; speaker/prosody: speaker similarity vs.\ energy contour) due to asymmetric bottlenecks in source variability and target synthesis.
Finally, human evaluation across dubbing, podcasts, and medical contexts shows that standalone MOS predictors can correlate \textit{negatively} with emotional fidelity, while top domain-specific metrics correlate strongly with human judgment (Spearman's $\rho \ge$ 0.90) and COMPASS rankings match human preference rankings across domains.

\vspace{1mm}
\noindent \textbf{Contributions.}
Our main contributions are:
(i) We introduce COMPASS, the first unified, modular framework for offline S2ST evaluation, integrating 40+ metrics across 8 axes;
(ii) We conduct a large evaluation study, covering 1,248 model-language configurations on FLEURS and CVSS;
(iii) We identify compact, data-driven metric subsets via correlation filtering, reducing 46 metrics to 10 per direction (with 3 axes differing across X$\to$EN and EN$\to$X), preserving rankings (Spearman's $\rho>$0.80) while cutting evaluation time by $\approx2.5\times$;
(iv) We run a human evaluation proving automatic metrics must be domain-aware;
and (v) We release the COMPASS toolkit to support fair, reproducible, and domain-aware S2ST evaluation.

\section{Related Work}

\vspace{1mm}
\noindent \textbf{S2ST Models.}
Early S2ST systems chained ASR, MT, and TTS in a cascaded pipeline \cite{lavie1997janus, nakamura2006atr}. Modern direct end-to-end models \cite{jia2019direct, translatotron2, nachmani2024translatotron} and large multilingual systems such as SeamlessM4T \cite{barrault2023seamlessm4t, barrault2023seamless} and the Qwen-Omni family \cite{xu2025qwen25omnitechnicalreport, xu2025qwen3} now unify speech and text translation in a single architecture. Cascaded systems remain competitive by combining specialized models like Whisper \cite{radford2023robust}, NLLB \cite{costa2022no}, Gemma \cite{kamath2025gemma}, and CosyVoice \cite{cosyvoice}. COMPASS evaluates both paradigms under a unified protocol.

\vspace{1mm}
\noindent \textbf{S2ST Evaluation Metrics.}
S2ST evaluation borrows from adjacent fields, typically selecting metrics most established or best correlated with human judgment. Translation quality is assessed via, among others, BLEU \cite{papineni2002bleu} and ChrF++ \cite{popovic2017chrf++}, alongside learned metrics such as COMET \cite{rei2020comet} and BLASER \cite{chen2023blaser}. Audio naturalness is most often measured with UTMOS \cite{saeki2022utmos} and NISQA \cite{mittag2021nisqa}, top VoiceMOS\footnote{\href{https://sites.google.com/view/voicemos-challenge}{\texttt{sites.google.com/view/voicemos-challenge}}} performers, though alternatives exist \cite{cooper2022generalization}. Speaker preservation typically uses embedding similarity from ECAPA-TDNN \cite{desplanques2020ecapa} or WavLM \cite{chen2022wavlm}, while prosody and emotion consistency rely on AutoPCP \cite{barrault2023seamless} and emotion recognition models \cite{ma2024emotion2vec}. Timing aspects (isochrony, isometry) have been studied mainly in dubbing \cite{lakew2022isometric, effendi2022duration, chronopoulou2023jointly}.

\vspace{1mm}
\noindent \textbf{Toward a Unified Evaluation.}
Adjacent fields have converged toward shared evaluation protocols, supported by standard corpora such as FLEURS \cite{conneau2023fleurs} and CVSS \cite{jia2022cvss} and domain-specific sets like MELD-ST \cite{chen2024meld} and MultiMed-ST \cite{le2025multimed}.
Text translation relies on standardized metric suites (BLEU, chrF, COMET) despite ongoing concerns about human correlation \cite{mathur2020tangled, freitag2021experts}. 
Speech-to-text translation (S2TT) has followed: recent IWSLT offline tracks adopted a shared set (COMET, BLEU, chrF, TER), with COMET as the primary ranking metric since 2024 \cite{ahmad-etal-2024-findings, agostinelli-etal-2025-findings}, while simultaneous S2TT has converged around unified streaming toolkits \cite{gaido2025simulstream}. 
Offline S2ST lacks a comparable protocol: state-of-the-art models report performance over heterogeneous, paper-specific axes; for instance, SeamlessM4T evaluates, among others, ASR-BLEU, BLASER, and AutoPCP \cite{barrault2023seamlessm4t}, TransVIP adds duration-based metrics \cite{le2024transvip}, and other systems mix or omit perceptual and timing dimensions \cite{cheng2025uniss, hu2025slm}, with further inconsistencies in metric implementations (e.g., ASR backends for ASR-BLEU). COMPASS fills this gap by unifying 46 metrics into a single toolkit.

\section{The COMPASS-S2ST Framework}
\label{sec:methodology}
COMPASS combines a structural evaluation taxonomy, a unified metric catalogue, and a data-driven filtering pipeline to select compact metric subsets.

\subsection{Design Goals}
COMPASS targets 4 core principles:
(i) \textit{multi-dimensionality}: capturing distinct quality axes since no single metric fits all applications;
(ii) \textit{modularity}: utilizing self-contained components so users can customize or extend specific metrics according to their use cases;
(iii) \textit{reproducibility}: providing a single toolkit with fixed checkpoints and deterministic configurations;
(iv) \textit{domain-awareness}: enabling domain-specific metric subsets validated by human judgment.

\subsection{\textit{A Priori} Taxonomy of S2ST Evaluation}
\label{sec:taxonomy}
We organize S2ST evaluation along eight axes, mapping to real-world application needs (Table~\ref{tab:domain_metrics}). 
These span core communicative requirements (translation, naturalness) alongside specialized context-driven constraints (e.g., timing for dubbing vs. prosodic richness for audiobooks).
\begin{enumerate}[noitemsep,topsep=5pt,partopsep=0pt]
    \item \textit{Translation Quality (Text)}: text-level adequacy and fluency of the translation, measured on the text output.\footnote{For end-to-end systems, we use the intermediate text produced alongside the synthesized speech, while for cascades, we use the MT text feeding into the TTS module. This ensures text-based metrics are applied consistently across both.}
    \item \textit{Translation Quality (ASR)}: translation quality measured on ASR transcripts of the synthesized audio, plus audio-grounded measures, e.g., BLASER and phoneme-level accuracy.
    \item \textit{Audio Naturalness}: perceptual quality of the synthesized speech.
    \item \textit{Speaker Consistency}: preservation of the source speaker identity.
    \item \textit{Prosody \& Emotion}: pitch, energy, rhythm, articulation, and emotional consistency.
    \item \textit{Isochrony}: temporal alignment between source and target speech.
    \item \textit{Isometry}: length compression and character or word ratios between source and target.
    \item \textit{Lipsync}: visual-acoustic alignment between source and target speech via viseme matching.
\end{enumerate}
These axes consolidate the evaluation dimensions adopted across recent S2ST systems and surveys \cite{barrault2023seamlessm4t, barrault2023seamless, le2024transvip, brannon2023dubbing}, jointly covering translation, perceptual, identity, prosodic, and timing aspects.
Empirically, our data collapse these into six non-redundant dimensions (Sec. \ref{sec:results}).
However, the original eight dimensions remain distinct, as the collapse reflects \textit{currently available} system profiles rather than redundancy of conceptual axes.

\begin{figure*}[t]
    \centering
    \includegraphics[width=1.7\columnwidth]{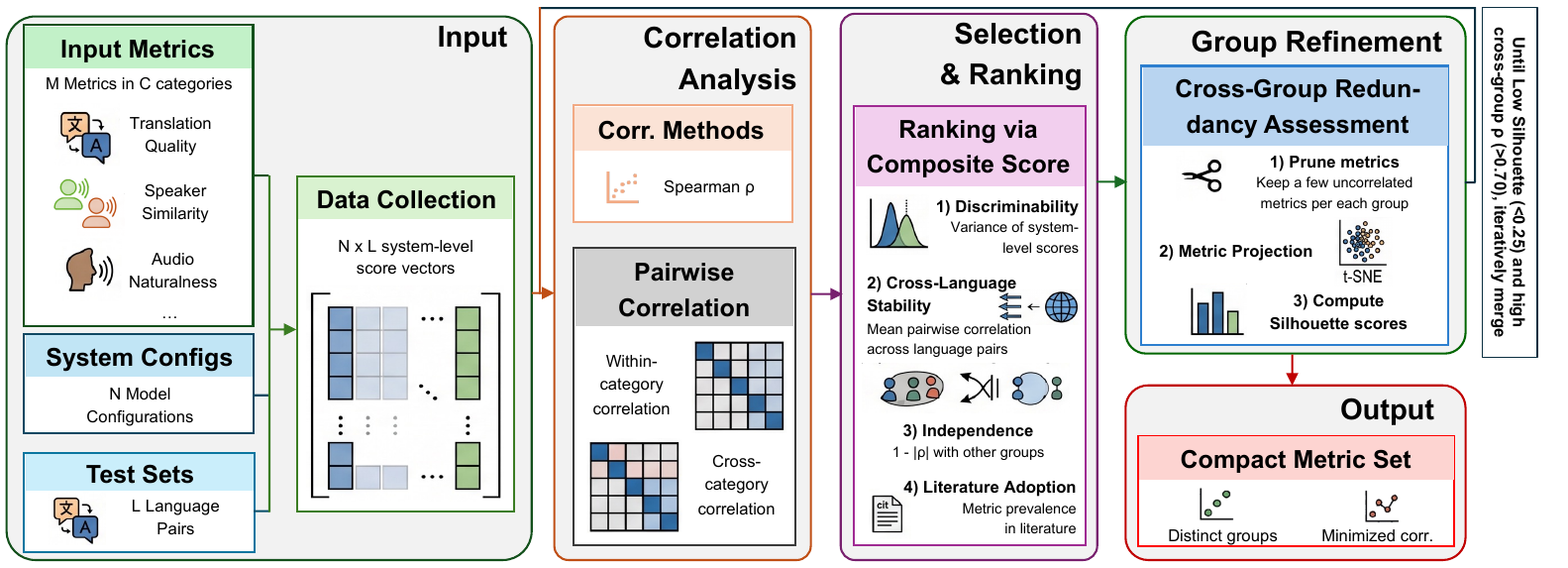}
    \caption{Data-driven filtering pipeline for COMPASS metrics.}
    \label{fig:filtering}
    \vspace{-3mm}
\end{figure*}
\subsection{Metric Catalogue}
\label{sec:metrics}
COMPASS integrates 46 metrics from established ASR, MT, TTS, and S2TT evaluation.
Translation quality is measured on both translated text and ASR transcripts via BLEU \cite{papineni2002bleu}, ChrF/ChrF++ \cite{popovic2015chrf, popovic2017chrf++}, TER \cite{snover-etal-2006-study}, COMET-DA and COMET-Kiwi \cite{rei-etal-2022-comet}, and Semantic Score \cite{phukon25_interspeech}. Speech translation quality is also assessed via BLASER \cite{chen2023blaser}, while WER measures intelligibility.
Audio naturalness uses UTMOS \cite{saeki2022utmos} and NISQA-MOS \cite{mittag2021nisqa}, top VoiceMOS predictors and de facto S2ST standards \cite{le2024transvip, cheng2025uniss}.
Speaker consistency is tracked via speaker embedding cosine similarity \cite{chen2022wavlm}.
Prosody and emotion span three sub-axes: (i) prosodic contours via AutoPCP \cite{barrault2023seamless}, F0/energy similarity, and speech-rate Spearman correlations over chars/syllables/words \cite{barrault2023seamless}; (ii) emotion via emotion match and embedding similarity \cite{ma2024emotion2vec}; and (iii) articulation via phoneme error rate, formant deviation, and amplitude envelope similarity.
Isochrony covers delta duration, relative duration error (RDE), speech overlap, duration ratio, and speech-length compliance \cite{brannon2023dubbing, le2024transvip}, capturing absolute, relative, and overlap-based alignment.
Isometry uses delta words/chars, char length ratio (LR), char compliance, and chars-per-second (CPS) ratio \cite{brannon2023dubbing, lakew2022isometric}, tracking text-length compression at surface and rate level.
Lipsync assesses video-audio alignment via viseme DTW and co-occurrence scores \cite{brannon2023dubbing}, focusing on phonetic mouth shapes.
Full metrics list and toolkits in Appendix~\ref{app:metrics}.

\subsection{From 46 Metrics to a Compact Set}
\label{sec:filtering}
To reduce metric redundancy, we implement a data-driven filtering pipeline, depicted in Fig.~\ref{fig:filtering}.

\vspace{1mm}
\noindent \textbf{Input.} For each metric $m$, we aggregate its utterance-level scores to the system level, yielding one vector of system scores per metric.

\vspace{1mm}
\noindent \textbf{Correlation Analysis.} We compute pairwise Spearman $|\rho|$ correlations between metrics, treating each as the vector of its system-level scores to capture how similarly two metrics rank systems. We convert correlation into distance matrices $D$ $= 1 - |\rho|$ and apply hierarchical clustering at threshold 0.15 ($|\rho| >$ 0.85) to group redundant metrics.

\vspace{1mm}
\noindent \textbf{Selection \& Ranking.} Within each cluster, we rank metrics on four complementary criteria:
\begin{itemize}[noitemsep,topsep=5pt,partopsep=0pt]
    \item \textit{Discriminability} ($D_m$): the coefficient of variation across systems; it measures how well the metric separates systems: low variance indicates poor discriminative power.
    \item \textit{Cross-language stability} ($C_m$): the mean Spearman $\rho$ of system rankings across language pairs; it captures ranking consistency regardless of the language evaluated.
    \item \textit{Independence} ($I_m$): defined as $1 - \overline{|\rho_m|}$, where $\overline{|\rho_m|}$ is the mean absolute correlation with all other metrics; higher values signify unique information content.
    \item \textit{Literature adoption} ($L_m$): a normalized usage score ($[0,1]$) based on evaluation frequency in published S2TT and S2ST papers; higher values reflect stronger community adoption.
\end{itemize}
Each component is min-max normalized to [0,1] and combined into a ranking score per metric:\footnote{$L_m$ introduces a \textit{popularity} bias, but as one of 4 equally-weighted components, it cannot dominate selection; it breaks ties in favor of comparability with prior work.}
\begingroup
\setlength{\abovedisplayskip}{6pt}
\setlength{\belowdisplayskip}{6pt}
\setlength{\abovedisplayshortskip}{0pt}
\setlength{\belowdisplayshortskip}{0pt}
\begin{equation}
    \mathrm{score}(m) = \frac{1}{4} \bigl( D_m + C_m + I_m + L_m \bigr)
    \label{eq:score}
\end{equation}%
\endgroup
A sensitivity analysis perturbing each weight by $\pm$20\% yielded stable rankings (Spearman's $\rho >$ 0.80), justifying the unweighted formula. The highest-scoring metrics per cluster become its representatives, with the count scaled to cluster size.

\vspace{1mm}
\noindent \textbf{Group Refinement \& Output.} We iteratively merge cross-dimension groups whose representatives remain correlated ($\rho_\text{cross} >$ 0.70 or silhouette $<$ 0.25), yielding a final compact set of non-redundant metrics organized into distinct empirical dimensions. Full pipeline details in Appendix~\ref{app:filtering}.

\vspace{1mm}
\noindent \textbf{Computational Cost.}
On an A6000 GPU, running the full metric suite takes nearly 10,845s ($\approx$3h) per configuration (CVSS), with neural metrics dominating.
The compact subsets reduce this cost by roughly 2.5$\times$ (4,300s) while preserving system rankings against the full 46-metric suite (Spearman's $\rho >$ 0.80 across both directions), making domain-aware evaluation practical for iterative development without sacrificing comparability.

\section{Experimental Setup}
\label{sec:experimental-setup}

\noindent \textbf{Datasets.}
We employ two widely used corpora:

\vspace{1mm}
\noindent \textit{FLEURS}~\cite{conneau2023fleurs} is a read-speech corpus.
We evaluate both translation directions (X$\leftrightarrow$EN) on the official test split across ten languages.: Arabic (\texttt{ar}), German (\texttt{de}), Spanish (\texttt{es}), French (\texttt{fr}), Hindi (\texttt{hi}), Italian (\texttt{it}), Japanese (\texttt{ja}), Korean (\texttt{ko}), Portuguese (\texttt{pt}), and Chinese (\texttt{zh}). These languages are jointly supported by the benchmarked systems and span different language families, covering both Latin and non-Latin scripts.

\vspace{1mm}
\noindent \textit{CVSS}~\cite{jia2022cvss} (built on CoVoST 2 \cite{wang2020covost}): we use it as a source-audio and reference-text resource (X$\to$EN),\footnote{We exclude EN$\to$X because synthetic English audio paired with target references would bias the evaluation.} as our metrics evaluate the synthesized output against the text reference.
We use 8 language pairs, excluding \texttt{hi} and \texttt{ko}, absent in CVSS.
For efficiency, metrics are computed on the first 1,000 utterances per pair.

\vspace{2mm}
\noindent \textbf{Systems.}
We benchmark end-to-end and cascaded architectures covering dominant paradigms: direct spanning dedicated S2ST models and multimodal speech LLMs, two-stage S2TT+TTS, three-stage ASR+MT+TTS. Full list of 1,248 system-language configurations is in Appendix~\ref{app:systems}.

\vspace{1mm}
\noindent \textit{End-to-end systems.}
We evaluate four models in standard inference mode: SeamlessM4T-Medium and Large-v2~\cite{barrault2023seamless}, Qwen2.5-Omni and Qwen3-Omni~\cite{xu2025qwen25omnitechnicalreport, xu2025qwen3}.

\vspace{1mm}
\noindent \textit{S2TT+TTS cascades.}
We pair eight S2TT models with two speech synthesis systems. 
For S2TT, we consider Whisper-Lv2 and Lv3~\cite{radford2023robust},\footnote{X$\to$EN only, since Whisper does not support EN$\to$X.} SeamlessM4T-Medium and Lv2, Gemma-4 E4B~\cite{gemma4}, Voxtral-Small~\cite{liu2025voxtral}, Qwen2.5-Omni, and Qwen3-Omni.
As TTS models, we use Chatterbox-TTS~\cite{chatterbox} and CosyVoice3~\cite{cosyvoice}.

\vspace{1mm}
\noindent \textit{ASR+MT+TTS cascades.}
We evaluate three-stage cascaded pipelines combining: ASR (Whisper-Lv2 and Lv3, SeamlessM4T-Medium and Lv2), MT (NLLB-200~\cite{costa2022no}, Gemma-3-27B~\cite{kamath2025gemma}, Qwen3-8B~\cite{yang2025qwen3}), and TTS (CosyVoice3 and Chatterbox).

\vspace{2mm}
\noindent \textbf{Evaluation Protocol.}
The pipeline transcribes target audios via Whisper-Lv3, computes all metrics, then averages to system-level (WER is inverted to Word Accuracy in visualizations, so higher is better).
All setups use identical inputs, fixed checkpoints, greedy decoding, and constant seeds, ensuring variance reflects genuine system behavior.

\vspace{1mm}
\noindent \textit{Reproducibility check.}  
We compared BLEU/ASR-BLEU for SeamlessM4T-Lv2 to the original report \cite{barrault2023seamless} to validate correctness.
Our scores align within $\pm$0.8 points across all languages (mean abs. difference: 0.3), with minimal variations attributable to normalization and ASR versioning, confirming absence of systematic bias.

\begin{table}[!t]
    \centering
    \setlength{\tabcolsep}{4pt}
    \renewcommand{\arraystretch}{0.96}
    \resizebox{\columnwidth}{!}{%
    \begin{tabular}{l l l}
        \toprule
        \textbf{Dir.} & \textbf{Dimension} & \textbf{Representatives} \\
        \midrule
        \multirow{6}{*}{\rotatebox{90}{X$\to$EN}}
         & Translation (ASR)           & \texttt{COMET-DA}, \texttt{WER} \\
         & Translation (Text)          & \texttt{COMET-DA}, \texttt{TER} \\
         & Audio Naturalness           & \texttt{UTMOS} \\
         & Speaker \& Prosody          & \texttt{Speaker} \texttt{Sim}, \texttt{SR} \texttt{Chars} \texttt{Spearman} \\
         & Isochrony                   & \texttt{RDE}, \texttt{Delta} \texttt{Duration} \\
         & Isometry                    & \texttt{Chars Length Compliance} \\
        \midrule
        \multirow{6}{*}{\rotatebox{90}{EN$\to$X}}
         & Translation (ASR)           & \texttt{COMET-DA}, \texttt{WER} \\
         & Translation (Text)          & \texttt{COMET-DA}, \texttt{ChrF++} \\
         & Audio Naturalness           & \texttt{NISQA-MOS} \\
         & Speaker \& Prosody          & \texttt{Energy} \texttt{Contour} \texttt{Sim}, \texttt{SR} \texttt{Chars} \texttt{Spearman} \\
         & Isochrony                   & \texttt{RDE}, \texttt{Delta Duration} \\
         & Isometry                    & \texttt{Chars} \texttt{Length} \texttt{Compliance} \\
        \bottomrule
    \end{tabular}}
    \caption{Compact metric subsets, X$\to$EN and EN$\to$X.}
    \label{tab:compact_metrics}
    \vspace{-4mm}
\end{table}

\section{Results}
\label{sec:results}
We organize the results across FLEURS and CVSS around four research questions (RQs):
\begin{description}[noitemsep, topsep=5pt,partopsep=0pt]
    \item[\textbf{RQ1:}] Which metrics are redundant, and which carry an independent signal?
    \item[\textbf{RQ2:}] Do X$\to$EN and EN$\to$X require the same evaluation metrics?
    \item[\textbf{RQ3:}] Does any single S2ST architecture dominate across all evaluation dimensions?
    \item[\textbf{RQ4:}] How do language proficiency profiles differ across architectures and language families?
\end{description}
To keep comparisons tractable while maximizing diversity, we select one representative per architectural family, i.e., the system with the highest mean rank across the six compact-set axes, computed per direction, with ties broken by literature adoption.\footnote{Findings are robust to this choice: re-running second-best systems preserves architectural ordering across all dimensions.}
Specifically, we evaluate one system per paradigm: Qwen3-Omni (\texttt{Q3O}, SpeechLLM), Seamless-Large v2 (\texttt{SeamL}, dedicated S2ST), Voxtral+Chatterbox (\texttt{S2TT+TTS}), and Whisper-Lv3+Gemma3+CosyVoice3 (\texttt{ASR+MT+TTS}).

\begin{figure*}[t]
    \centering
    \includegraphics[width=1.9\columnwidth]{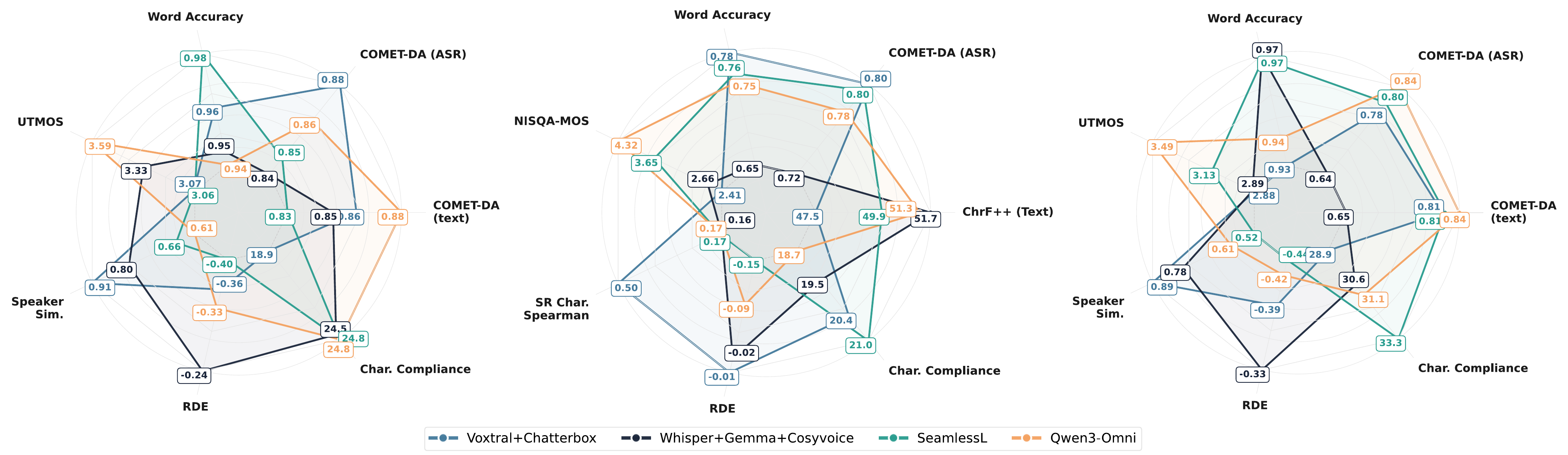}
    \caption{Models' performance on key metrics. FLEURS X$\rightarrow$EN (left), EN$\rightarrow$X (middle), CVSS X$\rightarrow$EN (right).}
    \label{fig:performance-radar-plots}
    \vspace{-3mm}
\end{figure*}

\vspace{2mm}
\noindent \textbf{RQ1: Metric Redundancy.}
The filtering pipeline (Sec. \ref{sec:filtering}) reduces the initial metrics to a compact set spanning six empirical dimensions (Table~\ref{tab:compact_metrics}).
In Appendix~\ref{app:compact_metrics}, we analyze a t-SNE projection in system-score space (Fig.~\ref{fig:subset-metrics-en-x-en}) and Spearman correlation matrices (Fig.~\ref{fig:spearman-rho-subsets}) to confirm these metrics form distinct, non-overlapping clusters.
Empirically, several original axes show strongly correlated scores across our benchmarked systems.
On our read-speech benchmarks, \textit{Speaker Consistency} and \textit{Prosody \& Emotion} merge into a single cluster due to high cross-system correlation, and \textit{Lipsync} is absorbed into \textit{Isochrony} for the same reason.
Conversely, \textit{Isochrony} and \textit{Isometry} remain distinct: their top representatives, RDE and Chars Compliance, correlate weakly ($\rho$ = -0.55 in EN$\to$X; $\rho$ = 0.26 in X$\to$EN), confirming speech and text alignment capture separate failure modes \cite{rozanov2024isochronometer}.
Most cross-dimension metric pairs maintain low correlations ($|\rho| <$ 0.50).
A control analysis on the spontaneous MELD-ST corpus weakens the speaker-prosody merge ($\rho \approx$ 0.75, Appendix~\ref{app:meldst_decoupling}), but the link remains strong.
This six-dimensional collapse likely reflects the joint variance profiles of \textit{current} S2ST architectures on predominantly read-speech benchmarks, rather than conceptual equivalence: read-speech corpora may suppress prosodic and emotional variability, inflating the correlation between speaker- and prosody-related metrics, while spontaneous-speech evaluation could decouple them further. 
Similarly, contemporary models excelling at voice cloning tend to preserve prosody and articulation simultaneously, so their errors vary jointly. We thus maintain the full 8-axis taxonomy as COMPASS's core conceptual framework, treating the 6-dimensional grouping as an empirical snapshot to be re-estimated as architectures and benchmarks evolve.

\vspace{2mm}
\noindent \textbf{RQ2: Direction-Specific Metric Subsets.}
Optimal metrics for X$\to$EN and EN$\to$X differ in composition and dependency structure, reflecting distinct bottlenecks (Table~\ref{tab:compact_metrics}).
X$\to$EN retains \textit{speaker similarity} while EN$\to$X preserves \textit{energy contour similarity}: source-side speaker variability dominates the former, target-side prosodic synthesis the latter.
\textit{Text evaluation} shifts from TER in X$\to$EN (tracking word-level edits in morphologically simple English) to ChrF++ in EN$\to$X, which better handles non-English morphology and CJK scripts \cite{popovic2017chrf++}.
\textit{Naturalness} moves from UTMOS (X$\to$EN) to NISQA-MOS (EN$\to$X), consistent with their training profiles: UTMOS is optimized for English \cite{saeki2022utmos}, whereas NISQA-MOS uses a multilingual mixture \cite{mittag2021nisqa}, generalizing better across target languages.
Internal correlations mirror this asymmetry.
In X$\to$EN, \textit{acoustic}, \textit{prosodic}, and \textit{timing} metrics form an entangled block, with $|\rho|$ in the $0.3$-$0.5$ range across WER, COMET-DA, NISQA-MOS, speech rate correlation, energy contour similarity, and $\Delta$ duration.
Conversely, in EN$\to$X, they decouple into two distinct clusters: one isolating \textit{text quality}, another linking \textit{naturalness}, \textit{speaker similarity}, and \textit{prosodic rate}.
This divergence implies that evaluating a single translation direction under-tests critical dimensions of the opposite path.

\vspace{2mm}
\noindent \textbf{RQ3: Architectural Trade-Offs.}
Fig.~\ref{fig:performance-radar-plots} confirms no single architecture dominates: cascades excel at timing and speaker preservation thanks to dedicated TTS, while end-to-end models produce more natural speech and avoid error propagation; translation quality is comparable across paradigms.
The trade-off is highly asymmetric: best-vs-worst gaps exceed 30\% on naturalness and speaker preservation but are minimal on translation quality. COMET-DA alone suggests rough equivalence, whereas voice or timing metrics yield radically different rankings, directly motivating COMPASS: \textit{single-metric S2ST evaluations are structurally misleading}, hiding the dimensions where architectural families actually diverge (further details in Appendix~\ref{app:rq3_full}).

\begin{figure*}[t]
    \centering
    \includegraphics[width=1.76\columnwidth]{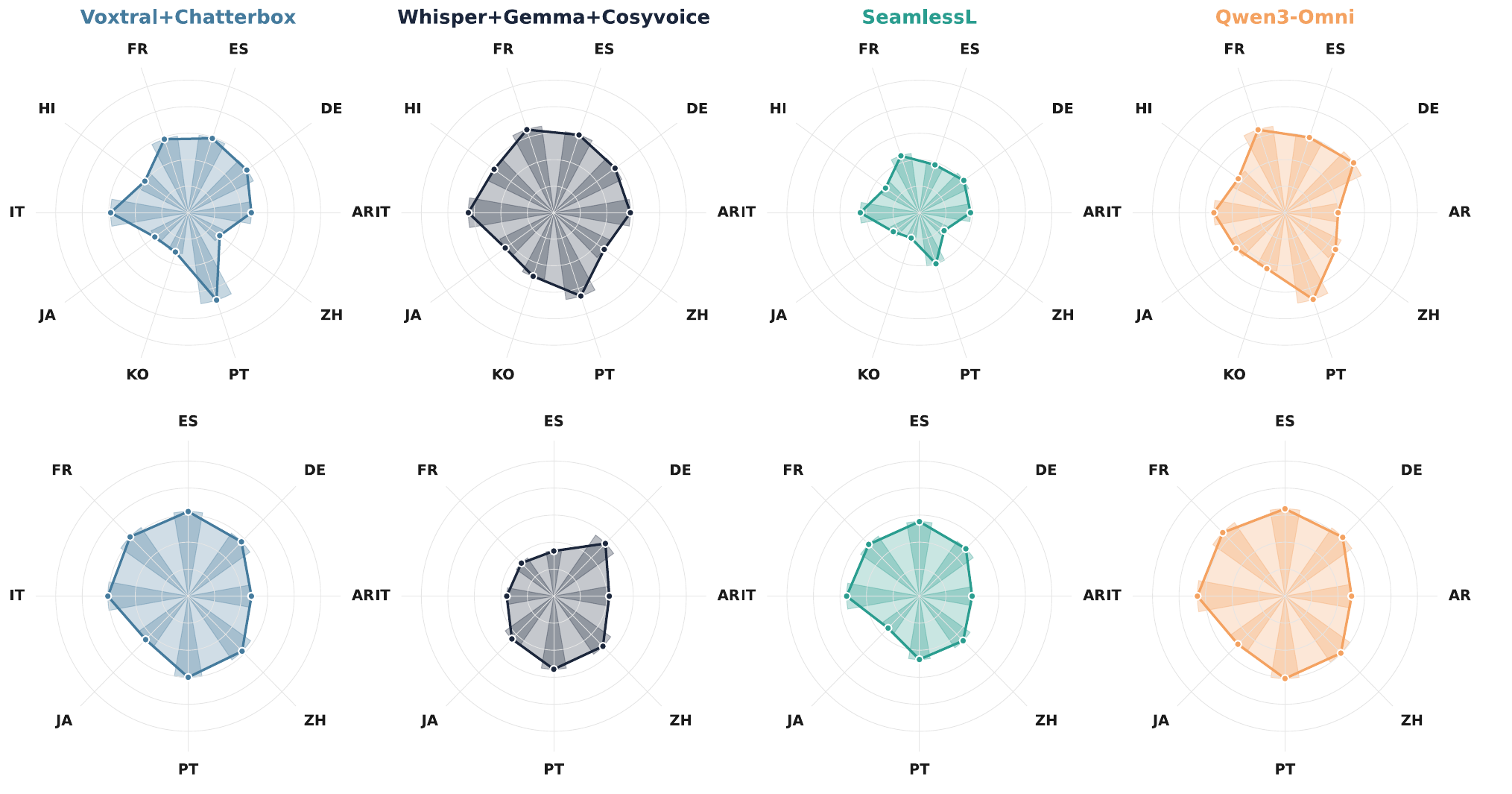}
    \caption{Language proficiency, FLEURS (top) and CVSS (bottom), X$\rightarrow$EN. 4 models, Voxtral+Chatterbox (\texttt{S2TT+TTS}), Whisper+Gemma+Cosyvoice (\texttt{ASR+MT+TTS}), Seamless (dedicated), and Qwen3Omni (SpeechLLM).}
    \label{fig:petal_xen}
    \vspace{-3mm}
\end{figure*}

\noindent \textbf{RQ4: Language Proficiency Profiles.}
We map each system's language proficiency by averaging normalized scores across six dimensions (Fig.~\ref{fig:petal_xen} X$\to$EN; EN$\to$X in Appendix~\ref{app:proficiency}, Fig.~\ref{fig:petal_enx}).
End-to-end models show more symmetric profiles across languages and directions than cascades, likely reflecting their jointly trained encoder-decoder distributing capacity across all language pairs.
On FLEURS X$\to$EN, \texttt{Q3O} remains stable across Germanic, Romance, and CJK languages, consistent with its broad multilingual pretraining.
Cascades exhibit severe directional asymmetry tied to component coverage: \texttt{ASR+MT+TTS} matches end-to-end models on FLEURS FR/IT$\to$EN but collapses on EN$\to$HI/KO, where TTS data is scarcer; this demonstrates performance is thus bounded by the weakest component per language.
On CVSS, \texttt{ASR+MT+TTS} also falls behind on Romance sources, suggesting its FLEURS advantage reflects corpus-specific conditions rather than genuine language coverage.
\texttt{SeamL} shows the opposite pattern: compressed on FLEURS X$\to$EN with the lowest CJK scores, yet competitive on EN$\to$X (e.g., DE, FR).\footnote{These proficiency differences are observational: disentangling training data, architecture, and benchmark coverage would require controlled ablations beyond our scope.}

\section{Human Evaluation}
\label{sec:human_eval}
To anchor COMPASS in real-world preferences, our human evaluation tests whether distinct domains require unique metric subsets and validates which metrics align best with human judgment.

\vspace{2mm}
\noindent \textbf{Domains.}
We evaluate three domains chosen to span the range of practical S2ST applications:
\begin{itemize}[noitemsep,topsep=5pt,partopsep=0pt]
    \item \textit{Dubbing} (MELD-ST~\cite{chen2024meld}, EN$\leftrightarrow$JA): Short, emotionally rich dialogue clips demanding precise timing, lip-sync, and prosody. We select 25 emotion-dense clips per direction (15-20s each; 50 total) using an automated difficulty metric (Appendix~\ref{app:human_eval_data}).   
    \item \textit{Podcasts} (EuroParl Multimedia Centre,\footnote{\url{https://multimedia.europarl.europa.eu}} EN$\leftrightarrow$\{DE, ES, FR, IT\}): long-form conversational speech prioritizing naturalness and content fidelity. We manually segment and extract six 30-second clips per direction per language pair (48 total).
    \item \textit{Medical} (MultiMed-ST~\cite{le2025multimed}, EN$\leftrightarrow$ZH): clinician-patient dialogues requiring exact accuracy. We select 20 clips per direction, single-speaker (8-14s each; 40 total), manually verified to contain medical terms.
\end{itemize}
Across all domains, we benchmark the four representative systems alongside ground-truth references where available (see Appendix~\ref{app:human_eval_data}).

\vspace{2mm}
\noindent \textbf{Annotation protocol.}
Each clip is evaluated by three independent native speakers of the target language with business-level English proficiency.
Annotators listen to source and synthesized target audios, rating them on a 5-point Likert scale across dimensions including translation quality, naturalness, speaker similarity, prosody, and domain-specific usability metrics\footnote{For medical dialogues, annotators are provided with reference text and a domain glossary to ensure accurate judgments.} (questionnaire in Appendix~\ref{app:human_eval_protocol}).

\vspace{2mm}
\noindent \textbf{Inter-Annotator Agreement (IAA).}
Agreement profiles vary by domain (Tables~\ref{tab:iaa_dubbing}-\ref{tab:iaa_medical} in Appendix~\ref{app:human_eval_iaa}).
Podcast ratings are highly reliable across all dimensions ($\alpha$ = 0.87-1.00), reaching near-perfect consensus on \textit{overall quality} ($\alpha$ = 0.98).
Medical annotations remain consistently robust ($\alpha$ = 0.76-0.89), anchoring strongest on \textit{real-world trust} ($\alpha$ = 0.89), lowest on \textit{prosody} ($\alpha$ = 0.76).
Conversely, dubbing shows lower baseline agreement ($\alpha$ = 0.68-0.85); while \textit{translation accuracy} and \textit{overall quality} remain highly stable ($\alpha \approx$ 0.85), human alignment drops significantly on highly granular temporal traits like \textit{lip sync} ($\alpha$ = 0.68) and \textit{timing} ($\alpha$ = 0.72).
This variance yields two key insights.
First, dimensions with the lowest human consensus are precisely where automated metrics are most valuable. 
Because tracking fine-grained audio timing is difficult for human judges to score consistently, automated metrics fill an important evaluation gap.
Second, these distinct agreement profiles empirically support the a priori taxonomy in Table~\ref{tab:domain_metrics}: human priorities vary across applications, confirming that evaluation configurations should shift with the deployment context.

\vspace{2mm}
\noindent \textbf{Metric-Human Alignment.}
We compute system-level Spearman $\rho$ between COMPASS scores and averaged human ratings, with bootstrap 95\% CIs over 1,000 resamples; correlations are reliable only when $|\rho|$'s lower bound exceeds 0.5.\footnote{Rankings are computed over four to five systems (MultiMed-ST does not have ground-truth target audios), so $\rho$ can saturate at $\pm 1.00$ for monotonic relationships; saturation indicates consistent ranking, not tight quantitative alignment.} Unless otherwise specified, reported values average across questions and directions within each domain; per-domain recommended subsets are in Appendix~\ref{app:human_eval_recommendations}.

\vspace{1mm}
\noindent \textit{Dominant metrics shift across domains.} 
Strongest predictors of human judgment vary by context.
Translation-quality metrics (\texttt{COMET-DA}, \texttt{Semantic} \texttt{Score}, \texttt{ChrF++}) lead in podcasts ($\rho$ = 0.82, CI: [0.75, 0.95]) and medical dialogues ($\rho$ = 1.00 for \texttt{COMET-DA} \texttt{(Text)} on EN$\to$ZH, CI: [0.72, 1.00]); prosody and timing metrics (\texttt{AutoPCP}, \texttt{CPS} \texttt{Ratio}, \texttt{$\Delta$} \texttt{Duration}) dominate dubbing ($\rho$ = 0.91, CI: [0.78, 0.99]; $\rho$ = -1.00 for \texttt{CPS Ratio} on EN$\to$JA, CI: [-1.00, -0.77]);
isometry metrics (\texttt{Chars} \texttt{Compliance}, \texttt{Chars} \texttt{LR}) are uniquely predictive of podcast naturalness and overall quality ($\rho \in$ [0.95, 1.00], CIs $\geq$ [0.72, 1.00]), suggesting excessive translation length disrupts long-form listening regardless of semantic accuracy.
This domain-conditional ordering broadly aligns with Table~\ref{tab:domain_metrics}'s priorities and confirms that no single metric subset transfers across domains: each application demands a distinct evaluation configuration.

\vspace{1mm}
\noindent \textit{MOS predictors mislead.}
\texttt{UTMOS} and \texttt{NISQA-MOS} yield near-zero or negative correlations across all domains; \texttt{NISQA-MOS} even correlates negatively with emotional preservation in dubbing ($\rho$ = -0.90, CI: [-1.00, -0.69]), suggesting clean synthesis can strip expressive prosody. MOS predictors appear to capture acoustic quality but fail to predict the higher-level dimensions listeners actually weigh.

\vspace{1mm}
\noindent \textit{Automatic metrics fill gaps where human consensus degrades.}
Domains with the lowest IAA (dubbing timing $\alpha$=0.72, lipsync $\alpha$=0.68) show the strongest automatic predictions ($\rho \geq 0.94$ for \texttt{$\Delta$} \texttt{Duration} and \texttt{Speech} \texttt{Overlap}). Conversely, dimensions where humans agree strongly but no metric reliably predicts, e.g., \textit{real-world trust} in medical ($\alpha$=0.89), motivate future specialized evaluation components.

\vspace{1mm}
\noindent \textit{Takeaways.} Across domains, \texttt{COMET-DA} and \texttt{Semantic} \texttt{Score} correlates the strongest with human judgment ($|\rho| \ge$ 0.79), while \texttt{UTMOS}/\texttt{NISQA-MOS} fail universally. 
Beyond this core, optimal metric subsets diverge by domain: prosody/timing for dubbing, isometry for podcasts, and text-level translation for medical, confirming that S2ST evaluation must be domain-conditional (top-10 metrics in Table~\ref{tab:global_top10}, Appendix~\ref{app:human_eval_corr}).

\vspace{2mm}
\noindent \textbf{System Rankings.}
Human evaluation confirms no single architecture dominates, and \textit{preference rankings align with COMPASS}: both place \texttt{SeamL} last (zero first-preference votes across all domains) and identify cascades as strongest on translation-critical tasks. 
In podcast X$\to$EN, \texttt{S2TT+TTS} matches the ground-truth reference (50\% each), the only instance where a system rivals human production; this edge disappears in EN$\to$X, where ground truth wins 89\% of clips. 
In dubbing, human tracks win across all clips, with EN$\to$JA scoring much lower than JA$\to$EN, likely reflecting limited Japanese paired training data rather than architectural limits. 
For medical dialogues, \texttt{ASR+MT+TTS} is preferred 70\% of the time in both directions, showing that a dedicated MT component outweighs end-to-end integration; \texttt{Q3O} ranks second (17\%), the only context where an end-to-end model beats a two-stage cascade (detailed rankings in Appendix~\ref{app:human_eval_rankings}).

\section{Conclusion}
\label{sec:conclusion}
We introduced COMPASS, the first framework for offline S2ST evaluation, combining 46 metrics across 8 axes. Evaluating 1,248 configurations reveals dimensional trade-offs across architectures, direction-specific metric requirements, and substantial redundancy in current evaluation practice.
Human evaluations confirm that effective metrics must adapt to deployment context, while MOS predictors fail to capture global preference.
Key gaps remain for future work: no automatic metric predicts real-world trust in medical contexts, dubbing timing shows low human consensus, and broader spontaneous-speech evaluation is needed to test whether the empirical 6-axes collapse persists as architectures and benchmarks evolve.


\section*{Limitations}
While COMPASS provides the largest empirical evaluation of offline Speech-to-Speech Translation to date, we acknowledge some limitations.

\vspace{1mm}
\noindent \textit{Read-speech bias in benchmarks.}
Both FLEURS and CVSS consist of read speech, which under-represents the prosodic and emotional variability of natural conversation and may weaken the discriminative power of prosody- and emotion-related metrics. Our human evaluation on MELD-ST (dubbing) and Multimedia EuroParl (podcasts) partly mitigates this by including spontaneous, emotionally rich material, but extending COMPASS to spontaneous-speech corpora such as VoxPopuli~\cite{wang2021voxpopuli} is a natural next step.

\vspace{1mm}
\noindent \textit{ASR self-evaluation loop.}
ASR-based metrics rely on \texttt{Whisper-Lv3}, the same model used as the front-end in several cascaded systems. This shared backbone may introduce an evaluation bias in favour of Whisper-based cascades. A control experiment (50 samples per language pair) with an alternative ASR backbone~\cite{pratap2024scaling} confirmed that relative rankings remain stable (Spearman's $\rho > 0.82$), so this bias does not affect our main conclusions.

\vspace{1mm}
\noindent \textit{Noisy source audio.}
Speaker and prosody metrics compare the synthesized output against the source audio directly. On data such as FLEURS, noisy source recordings can penalize systems that correctly denoise the input, as the clean synthetic embedding drifts from the noisy reference embedding. COMPASS does not currently apply source-side speech enhancement before metric computation, highlighting an open challenge in decoupling speech enhancement quality from translation fidelity during evaluation.

\vspace{1mm}
\noindent \textit{Offline evaluation only.}
COMPASS benchmarks fully formed utterances and does not track latency-related metrics such as Average Lagging and related scores \cite{ma2019stacl, ma2020simuleval, papi2022over}. End-to-end systems are evaluated on equal footing with cascaded models, hiding real-time deployment and inference costs trade-offs.

\vspace{1mm}
\noindent \textit{Sentence-level human evaluation.}
Our human annotators evaluated short clips of up to 30 seconds. Long-form phenomena such as speaker voice drift, cumulative timing delays, or progressive hallucination over longer audios are not captured by the current evaluation design.

\vspace{1mm}
\noindent \textit{EN$\to$X grounded in a single corpus.}
Our EN$\rightarrow$X analysis is grounded in FLEURS alone, since CVSS contains only synthetic English source audio in this direction. Extending COMPASS to additional EN$\rightarrow$X corpora is a natural next step to confirm the generality of our direction-specific recommendations.

\vspace{1mm}
\noindent \textit{Empirical grouping is system-dependent.}
The six-dimensional empirical grouping reflects the current generation of S2ST systems and may shift as architectures evolve, particularly if future systems decouple speaker, prosody, and articulation. We therefore release both the eight-dimensional taxonomy and the filtering pipeline, rather than a fixed metric set, so that the empirical grouping can be re-estimated as the field evolves.

\vspace{1mm}
\noindent \textit{GPU non-determinism.}
Some neural metrics (e.g., COMET, BLASER, WavLM-based speaker similarity) rely on GPU atomic operations whose accumulation order is non-deterministic, introducing run-to-run variance on the order of 1e-3. This noise is below the between-system score gaps reported in our analysis and does not alter system rankings.
 
\section*{Ethics Statement}

\vspace{1mm}
\noindent \textit{Annotator compensation and conditions.}
Human annotators were recruited through two channels: professional annotators compensated above the local minimum wage for their time, and voluntary researcher annotators with relevant domain expertise. Annotators were informed of the purpose of the study and provided consent before participating. No personally identifiable information was collected during the annotation process.

\vspace{1mm}
\noindent \textit{Licensing and data usage.}
All datasets used in this work (FLEURS, CVSS, MELD-ST, MultiMed-ST) are publicly available and used in accordance with their respective licenses. Podcast evaluation audio was downloaded from the official European Parliament Multimedia Centre and used in accordance with its terms of use. The medical evaluation data (MultiMed-ST) contains clinician-patient dialogues: we use only the portions released for research purposes and do not attempt to re-identify any individuals. The benchmark code will be released under a commercial license upon acceptance. Note that licenses for the underlying metrics and pretrained checkpoints vary by source and may impose additional restrictions; we detail each in Appendix~\ref{app:metrics}.

\vspace{1mm}
\noindent \textit{Dual-use and potential risks.}
COMPASS is an evaluation framework, not a generative system; we release no fine-tuned models and benchmark only publicly available systems. Nonetheless, the S2ST models evaluated here could be misused for voice cloning, unauthorized dubbing, or misleading audio content. COMPASS itself may also be misinterpreted if applied outside the tested conditions: metric recommendations are grounded in our benchmarked systems, languages, and domains, and transferring them to unsupported languages, low-resource settings, or specialized domains (e.g., legal, emergency) without re-validation may yield misleading rankings. In high-stakes contexts such as medical S2ST, our results show that no automatic metric reliably predicts real-world trust; COMPASS scores must not substitute for expert human review before deployment. We encourage users to treat COMPASS as a diagnostic tool that complements, rather than replaces, human judgment, and to consider the ethical implications of deploying S2ST systems in sensitive domains.

\vspace{1mm}
\noindent \textit{Societal impact.}
Reliable evaluation frameworks are a prerequisite for the responsible deployment of S2ST technology.
By identifying which metrics correlate with human judgment and which do not, COMPASS reduces the risk of systems being deployed based on misleading automatic scores. We hope this work contributes to more transparent and accountable evaluation practices in the speech translation community.

\vspace{1mm}
\noindent \textit{Ethics review.}
This research, including the human evaluation protocol and data collection procedures, was reviewed and approved by our institution's internal AI  Ethics board.

\vspace{1mm}
\noindent \textit{Usage of AI assistants.}
We used AI assistants for proofreading, table formatting, and assistance 
with data preparation scripts. All scientific content, experimental design, 
analysis, and conclusions are the authors' own.


\bibliography{custom}


\newpage

\appendix
\label{sec:appendix}

\section{The COMPASS Toolkit}
\label{app:compass-toolkit}
The COMPASS toolkit is organized around a metric registry that maps each metric to its implementation, expected inputs, and directionality (higher- or lower-is-better).
The pipeline runs end-to-end from raw audio inputs to per-system, per-language, and per-dimension scores, with all intermediate artefacts cached for reproducibility.
Adding a new metric requires only a single registry entry and a compute function, making the framework easy to extend as the field evolves.

\paragraph{Text pre-processing.} Before computing translation metrics, all hypothesis and reference texts are passed through a common normalization step that removes redundant whitespace, normalizes Unicode punctuation, and applies language-specific lower-casing where appropriate.\footnote{We rely on Whisper normalizers, available under MIT license: \href{https://github.com/openai/whisper/tree/main/whisper/normalizers}{\texttt{github.com/openai/whisper/normalizers}}.} The same normalization is applied uniformly across all translation quality metrics.

\paragraph{Hardware.} We run the entire evaluation benchmark on a single NVIDIA A6000 GPU (48GB) 
under Linux (Ubuntu 24.04, kernel 6.17, x86\_64). This also includes neural metrics (Whisper-Lv3 \cite{radford2023robust}, COMET \cite{rei2020comet, rei-etal-2022-comet}, BLASER \cite{chen2023blaser}, UTMOS-v2 \cite{saeki2022utmos}, NISQA-MOS \cite{mittag2021nisqa}, WavLM-SV \cite{chen2022wavlm}, AutoPCP \cite{barrault2023seamless}, Wav2Vec2 phoneme \cite{baevski2020wav2vec}, Emotion2Vec base plus \cite{ma2024emotion2vec}).

\paragraph{Timing Analysis.}
To evaluate operational throughput, we benchmarked computation times on a fixed subset of 100 samples across the IT$\to$EN translation direction.\footnote{Timing profiles were collected by benchmarking the first 100 consecutive utterances from each evaluation dataset.} For the FLEURS \cite{conneau2023fleurs} dataset, running the full metric suite required 1,597 seconds ($\approx$0.45h), whereas the non-redundant compact set completed execution in 689 seconds (2.3$\times$ reduction). Similarly, on the CVSS dataset \cite{jia2022cvss}, computing the full evaluation suite took 1,085 seconds, which dropped down to 430 seconds when utilizing the compact subset. 
This substantial reduction in compute footprint demonstrates that domain-specific metric configurations make iterative system development far more practical.

\section{Detailed Metric Catalogue}
\label{app:metrics}
This section provides details on the full set of 46 metrics integrated into COMPASS, organized by the eight \textit{a priori} evaluation dimensions described in Section~\ref{sec:taxonomy}.
For each metric, we report what it measures, the toolkit used, and the underlying model checkpoint where applicable.
All metrics are computed with deterministic configurations (greedy decoding, fixed sampling rate of 16~kHz) to ensure reproducibility.

\subsection{Translation Quality (Text)}
\label{app:tqtext}
These metrics evaluate the textual translation produced by the system. All texts are normalized through a common pre-processing step before scoring (see Appendix~\ref{app:compass-toolkit}).
\begin{itemize}[noitemsep,topsep=5pt,partopsep=0pt]
    \item \textit{BLEU}~\cite{papineni2002bleu}: corpus- and sentence-level n-gram precision with brevity penalty. Computed with \texttt{SacreBLEU}\footnote{SacreBLEU (\href{https://github.com/mjpost/sacrebleu}{\texttt{github.com/mjpost/sacrebleu}}) comes with Apache-2.0 license.}~\cite{post-2018-call} using \texttt{tokenize=None}, which auto-selects the tokenizer based on the target language: \texttt{zh} for Chinese, \texttt{ja-mecab} for Japanese, \texttt{ko-mecab} for Korean, and \texttt{13a} (mteval) otherwise.
    \item \textit{ChrF}~\cite{popovic2015chrf}: character-level F-score measuring character n-gram overlap up to length 6. Computed with \texttt{sacreBLEU} (\texttt{word\_order=0}) using a default recall-biased $\beta=2$.
    \item \textit{ChrF++}~\cite{popovic2017chrf++}: character-level F-score augmented with word n-grams up to length 2 to incorporate word-order constraints. Computed with \texttt{sacreBLEU} (\texttt{word\_order=2}).
    \item \textit{TER}~\cite{snover-etal-2006-study}: translation edit rate calculating the minimum edits (insertions, deletions, substitutions, block shifts) normalized by reference length. Computed with \texttt{sacreBLEU}'s \texttt{TER} class.
    \item \textit{COMET-DA}~\cite{rei2020comet}: reference-based learned metric using a XLM-RoBERTa~\cite{conneau2020unsupervised} encoder fine-tuned on human Direct Assessments. We use the official \texttt{wmt22-comet-da} model checkpoint.\footnote{\href{https://huggingface.co/Unbabel/wmt22-comet-da}{\texttt{huggingface.co/Unbabel/wmt22-comet-da}}, which has Apache 2.0 license.}
    \item \textit{COMET-Kiwi}~\cite{rei-etal-2022-comet}: reference-free quality estimation metric predicting human quality scores directly from source and hypothesis embeddings. We use the \texttt{wmt22-cometkiwi-da} checkpoint.\footnote{\href{https://huggingface.co/Unbabel/wmt22-cometkiwi-da}{\texttt{huggingface.co/Unbabel/wmt22-cometkiwi-da}}, having CC-BY-NC-SA license.}
    \item \textit{SemScore}~\cite{phukon25_interspeech}: weighted combination of NLI entailment\footnote{\href{https://huggingface.co/ynie/roberta-large-snli_mnli_fever_anli_R1_R2_R3-nli}{\texttt{huggingface.co/ynie/roberta-large-nli}}, MIT license} BERTScore, and phonetic similarity. BERTScore~\cite{bertscore} uses the language-appropriate backbone with rescaled baseline; phonetic similarity uses Soundex + Jaro-Winkler via \texttt{jellyfish}.\footnote{\href{https://github.com/jamesturk/jellyfish}{github.com/jamesturk/jellyfish}, MIT license} Default weights are $0.4012$~(NLI), $0.2785$~(BERTScore), $0.3201$~(phonetic). For non-English target languages, the phonetic weight is redistributed to NLI and BERTScore proportionally.
\end{itemize}

\subsection{Translation Quality (ASR)}
\label{app:tqasr}
These metrics apply text-based translation metrics to ASR transcripts of the synthesized audio, plus audio-grounded measures.

\paragraph{Reference ASR system.}
All following ASR-based metrics rely on transcripts produced by Whisper-large-v3,\footnote{\href{https://huggingface.co/openai/whisper-large-v3}{\texttt{huggingface.co/openai/whisper-large-v3}}, Apache 2.0 license} with language-conditioned generation and default decoding.
\begin{itemize}[noitemsep,topsep=5pt,partopsep=0pt]
    \item \textit{ASR-BLEU, ASR-ChrF, ASR-ChrF++, ASR-TER}: identical configuration to their text counterparts (Appendix~\ref{app:tqtext}), evaluated by swapping the model text output with its Whisper ASR transcript.
    \item \textit{COMET-DA (ASR), COMET-Kiwi (ASR), SemScore (ASR)}: text-based embedding and neural metrics applied directly to ASR transcripts to evaluate quality without using ground-truth source text scripts.
    \item \textit{Corpus WER}: word error rate of ASR transcripts against reference translations, computed with \texttt{jiwer},\footnote{\href{https://github.com/jitsi/jiwer}{\texttt{github.com/jitsi/jiwer}}, Apache 2.0 license} after the same text normalization used for BLEU. Empty-reference pairs are filtered out. For CJK languages, we use Character Error Rate (CER) in the EN$\rightarrow$X translation.
\end{itemize}

\paragraph{Audio grounded.}
These metrics evaluate translation quality directly from the raw acoustic waveforms without intermediate text transcription or automatic speech recognition.
\begin{itemize}[noitemsep,topsep=5pt,partopsep=0pt]
    \item \textit{BLASER 2.0}~\cite{chen2023blaser}: text-free, audio-grounded translation similarity. Computed as $1 - \cos(\mathbf{e}_{src}, \mathbf{e}_{tgt})$, where embeddings are obtained from language-specific SONAR speech encoders.\footnote{\href{https://github.com/facebookresearch/SONAR}{\texttt{github.com/facebookresearch/SONAR}}, with license either MIT or non-commercial depending on the specific language}
    \item \textit{Phoneme F1}: token-level F1 of phoneme bag overlap between source and target audio. Phonemes are extracted with a wav2vec 2.0 model~\cite{baevski2020wav2vec} fine-tuned to recognize phonetic labels in multiple languages\footnote{\label{footnote-w2v2-lv60}\href{https://huggingface.co/facebook/wav2vec2-lv-60-espeak-cv-ft}{\texttt{huggingface.co/facebook/wav2vec2-lv60-espeak}}, Apache 2.0 license} (greedy CTC decoding).
    \item \textit{Phoneme Distribution Similarity}: defined as $1 - \mathrm{JSD}(p_{src}, p_{tgt})$, where $p_{src}$ and $p_{tgt}$ are normalized phoneme histograms over the union vocabulary, and JSD is the Jensen-Shannon divergence~\cite{lin1991divergence}.
\end{itemize}

\subsection{Audio Naturalness}
\label{app:naturalness}
These metrics assess the perceptual quality and intelligibility of the synthesized speech, independently of the source.
\begin{itemize}[noitemsep,topsep=5pt,partopsep=0pt]
    \item \textit{UTMOS-v2}~\cite{saeki2022utmos}: neural mean opinion score predictor. It estimates synthesized speech naturalness by extracting acoustic features via an SSL backbone fine-tuned on diverse human ratings.\footnote{\href{https://github.com/sarulab-speech/UTMOSv2}{github.com/sarulab-speech/UTMOSv2}, MIT license} Score range: $[1, 5]$.
    \item \textit{NISQA-MOS}~\cite{mittag2021nisqa}: non-intrusive speech quality predictor that models multidimensional quality profiles including distortion, noisiness, and color. We invoke the official NISQA repository\footnote{\url{https://github.com/gabrielmittag/NISQA}, code released under MIT license, models released under CC BY-NC-SA 4.0} with the \texttt{nisqa\_tts.tar} TTS-tuned weights. Score range: $[1, 5]$.
    \item \textit{Intelligibility PPL}: token-level perplexity of phoneme posteriors from the same wav2vec 2.0 model fine-tuned on multilingual phonetic labels introduced above, defined as $\exp(\overline{H(p_t)})$, where $H(p_t)$ is the per-frame entropy of the CTC distribution. Lower values indicate higher structural intelligibility and clearer acoustic articulation through more confident, sharp phoneme predictions.
\end{itemize}

\subsection{Speaker Consistency}
\label{app:speaker}
This dimension measures how well the synthesized speech preserves the identity of the source speaker.
\begin{itemize}[noitemsep,topsep=5pt,partopsep=0pt]
    \item \textit{Speaker Similarity}: cosine similarity between L2-normalized x-vector embeddings of source and synthesized audio, extracted with the WavLM model~\cite{chen2022wavlm},\footnote{\href{https://huggingface.co/microsoft/wavlm-base-plus-sv}{\texttt{huggingface.co/microsoft/wavlm-base-plus-sv}}, MIT license} fine-tuned for speaker verification.
\end{itemize}

\subsection{Prosody \& Emotion}
\label{app:prosody}

\paragraph{Global prosody.}
These metrics compare prosodic properties of the source and synthesized speech at the utterance level, capturing pitch, energy, rhythm, and tempo agreement.
\begin{itemize}[noitemsep,topsep=5pt,partopsep=0pt]
    \item \textit{AutoPCP}~\cite{barrault2023seamless}: prosody comparator that scores how closely the prosody of the synthesized target speech matches the source speech. We use the official Meta implementation released as part of the \texttt{stopes} library.\footnote{\href{https://github.com/facebookresearch/stopes}{\texttt{github.com/facebookresearch/stopes}}, under MIT license (although some code and models follow CC BY-NC 4.0}
    \item \textit{F0 Contour Similarity}: Pearson correlation between source and target F0 contours, after linear interpolation to a common length. F0 is extracted with the \texttt{librosa} library\footnote{\href{https://librosa.org/doc/0.11.0/generated/librosa.pyin.html}{\texttt{librosa.org/librosa.pyin}}, licensed under the permissive ISC license} (default settings).
    \item \textit{Energy Contour Similarity}: Pearson correlation between RMS energy contours computed via \texttt{librosa},\footnote{\href{https://librosa.org/doc/0.11.0/generated/librosa.feature.rms.html}{\texttt{librosa.org/librosa.feature.rms}}} after linear interpolation.
    \item \textit{Rhythm Similarity}: Pearson correlation between onset-strength envelopes computed via \texttt{librosa}.\footnote{\href{https://librosa.org/doc/0.11.0/generated/librosa.onset.onset_strength.html}{\texttt{librosa.org/librosa.onset.onset\_strength}}}
    \item \textit{Tempo Ratio}: $\min(t_{src}, t_{tgt}) / \max(t_{src}, t_{tgt})$, where tempos are estimated with \texttt{librosa}.
\end{itemize}

\paragraph{Local prosody.} All local prosody metrics use Meta's \texttt{stopes} \texttt{eval.local\_prosody} toolkit~\cite{barrault2023seamless}.
\begin{itemize}[noitemsep,topsep=5pt,partopsep=0pt]
    \item \textit{Forced alignment.} We use the official forced aligner released within the fairseq library\footnote{\texttt{fairseq2\_nar\_t2u\_aligner}} to produce word-level timestamps.
    \item \textit{VAD refinement.} Pause boundaries are refined with the \texttt{stopes} built-in VAD (window size 512), which produces per-frame speech probabilities used to refine inter-word silence durations.
    \item \textit{Speaking Rate (SR Word / Syllable / Char):} number of words, syllables, or characters per second of net speech (excluding pauses). The corpus-level metrics \textit{SR Word/Syllable/Char Spearman} are the Spearman rank correlations between source and target speaking rates over all utterances in a system, capturing whether the model preserves \emph{relative} speaking-rate variation across the corpus.
    \item \textit{LP Mean Joint Score}: per-utterance product of three normalized agreement signals between source and target: (i) speaking-rate ratio agreement, (ii) pause-position agreement after dynamic time warping of word boundaries, and (iii) pause-duration agreement. The score is averaged across utterances.
    \item \textit{LP Weighted Joint Score}: same as LP Mean Joint Score, but utterances are weighted by source duration, giving more importance to longer utterances.
\end{itemize}

\paragraph{Emotion.}
These metrics measure how well the emotional content of the source speech is preserved in the synthesized output. All metrics employ the emotion2vec model \cite{ma2024emotion2vec},\footnote{\href{https://huggingface.co/emotion2vec/emotion2vec_plus_base}{\texttt{huggingface.co/emotion2vec\_plus\_base}}, licensed under MIT license} with the standard 8-class label set (neutral, calm, happy, sad, angry, fearful, disgust, surprised).
\begin{itemize}[noitemsep,topsep=5pt,partopsep=0pt]
    \item \textit{Emotion Match}: agreement between top-1 emotion labels predicted on source and synthesized audio.
    \item \textit{Emotion Probability Similarity}: it is computed as $1 - \mathrm{JSD}(p_{src}, p_{tgt})$ between source and target emotion probability distributions over the same 8 classes.
    \item \textit{Emotion Embedding Similarity}: cosine similarity between hidden-state embeddings of the same Emotion2Vec classifier, mean-pooled across the time dimension.
\end{itemize}

\paragraph{Articulation.}
These metrics compare the fine-grained acoustic-phonetic properties of source and synthesized speech, capturing pronunciation accuracy and articulatory similarity.
\begin{itemize}[noitemsep,topsep=5pt,partopsep=0pt]
    \item \textit{Phoneme Error Rate (PER)}: edit distance between phoneme sequences of source and target audio, normalized by source phoneme length. The phonemes are extracted with the wav2vec 2.0 model fine-tuned for the phoneme recognition task, with greedy CTC.
    \item \textit{Formant Deviation}: mean absolute deviation between F1, F2, and F3 formants of source and target, extracted with \texttt{Parselmouth},\footnote{\href{https://github.com/YannickJadoul/Parselmouth}{\texttt{github.com/YannickJadoul/Parselmouth}}, GPL-3.0 license.} a Python interface to Praat~\cite{boersma2001}. Formants are computed with the Burg method (max 5 formants, 5500~Hz ceiling).
    \item \textit{Amplitude Envelope Similarity}: Pearson correlation between Hilbert-transform amplitude envelopes of source and synthesized audio, after linear interpolation to a common length.
\end{itemize}

\subsection{Isochrony}
\label{app:isochrony}
These metrics measure the temporal alignment between source and synthesized speech, capturing duration agreement and speech-length compliance.
\begin{itemize}[noitemsep,topsep=5pt,partopsep=0pt]
    \item \textit{Delta Duration}: $|d_{tgt} - d_{src}|$ in seconds. It captures the absolute discrepancy between target and source audio lengths, penalizing over-generation or truncations regardless of input scale. Lower is better.
    \item \textit{RDE (Relative Duration Error)}: $|d_{tgt} - d_{src}| / d_{src}$, the source-normalized duration error. This scales the absolute duration difference relative to the input length, allowing uniform alignment error comparisons across both short utterances and long speech segments.
    \item \textit{Speech Overlap}: temporal overlap (in seconds) between source and target speech regions after voice activity detection (VAD), normalized by source duration~\cite{le2024transvip}. VAD is performed with \texttt{silero-vad}.\footnote{\href{https://github.com/snakers4/silero-vad}{\texttt{github.com/snakers4/silero-vad}}, MIT license.}
    \item \textit{Duration Ratio}: $d_{tgt} / d_{src}$, the raw duration ratio. It quantifies expansion or compression behaviors globally, where values above 1.0 indicate systematic target elongation and values below 1.0 reveal speech compression trends.
    \item \textit{SLC 0.2 / SLC 0.4 (\%)}: speech-length compliance~\cite{lakew2022isometric, le2024transvip}, the percentage of utterances whose duration ratio falls within $\pm 20\%$ or $\pm 40\%$ of the source duration.
\end{itemize}

\subsection{Lip-Sync}
These metrics estimate the visual-acoustic alignment between source and synthesized speech, capturing how well the lip movements implied by the target audio would match those of the source. Because our evaluation framework operates strictly on acoustic data, we model these traits exclusively at the phonetic level by mapping speech to visemes (the visual equivalents of phonemes). We use viseme tables\footnote{\href{https://docs.aws.amazon.com/polly/latest/dg/viseme.html}{\texttt{docs.aws.amazon.com/polly/latest/dg/viseme}}} to map each aligned phone to its corresponding viseme set, following the procedure in \cite{brannon2023dubbing}.
\begin{itemize}[noitemsep,topsep=5pt,partopsep=0pt]
    \item \textit{Lipsync Viseme DTW}: dynamic time warping similarity between viseme sequences extracted from source and synthesized audio. 
    \item \textit{Lipsync Co-occurrence}: fraction of time-aligned frames in which source and target visemes match, after frame-level alignment of the phoneme posterior streams \cite{brannon2023dubbing}.
\end{itemize}

\subsection{Isometry \& Length}
\label{app:isometry}
These metrics compare the textual length of source and target translations, capturing whether the target output stays within plausible length and speaking-rate ranges relative to the source.
\begin{itemize}[noitemsep,topsep=5pt,partopsep=0pt]
    \item \textit{Delta Chars}: $|c_{tgt} - c_{src}|$, the absolute difference in character counts between source and target text. 
    \item \textit{Delta Words}:\footnote{We found it to be correlated almost perfectly ($|\rho| > 0.95$) with Delta Chars across all language groups while being highly unreliable for languages without explicit word boundaries like CJK.} $|w_{tgt} - w_{src}|$, the absolute difference in word counts between source and target text strings, used primarily for whitespace-tokenized languages.
    \item \textit{Char Length Ratio}: $c_{tgt} / c_{src}$, quantifying the global structural expansion or contraction of the translation string.
    \item \textit{Char Compliant (\%)}: percentage of utterances whose character length ratio falls within an acceptable range relative to the source text~\cite{anastasopoulos2022findings, lakew2022isometric}.
    \item \textit{CPS Ratio}: ratio of characters-per-second between target and source, capturing whether the target speaking rate is realistic given the translation length.
\end{itemize}

\begin{table*}[!t]
    \centering
    \resizebox{2\columnwidth}{!}{%
    \begin{tabular}{llllc}
    \toprule
    \textbf{Type} & \textbf{System} & \textbf{Checkpoint} & \textbf{Mode} & \textbf{Directions} \\
    \midrule
    \multirow{6}{*}{\rotatebox{90}{End-to-End}}
     & SeamlessM4T-Medium                       & \texttt{seamless-m4t-medium}                                                                             & E2E S2ST                    & X$\leftrightarrow$EN \\
     & SeamlessM4T-Large-v2                     & \texttt{seamless-m4t-v2-large}                                                                           & E2E S2ST                    & X$\leftrightarrow$EN \\
     & Qwen2.5-Omni                             & \texttt{Qwen2.5-Omni-7B}                                                                                 & Standard                    & X$\leftrightarrow$EN \\
     & Qwen2.5-Omni                             & \texttt{Qwen2.5-Omni-7B}                                                                                 & CoM           & X$\leftrightarrow$EN \\
     & Qwen3-Omni                               & \texttt{Qwen3-Omni-30B-A3B-Instruct}                                                                     & Standard                    & X$\leftrightarrow$EN \\
     & Qwen3-Omni                               & \texttt{Qwen3-Omni-30B-A3B-Instruct}                                                                     & CoM           & X$\leftrightarrow$EN \\
    \midrule
    \multirow{16}{*}{\rotatebox{90}{S2TT + TTS}}
     & Whisper-Large-v2     + CosyVoice3        & \texttt{whisper-large-v2}            + \texttt{Fun-CosyVoice3-0.5B-2512}                                 & Cascade                     & X$\to$EN \\
     & Whisper-Large-v2     + Chatterbox        & \texttt{whisper-large-v2}            + \texttt{ChatterboxMultilingualTTS}                                & Cascade                     & X$\to$EN \\
     & Whisper-Large-v3     + CosyVoice3        & \texttt{whisper-large-v3}            + \texttt{Fun-CosyVoice3-0.5B-2512}                                 & Cascade                     & X$\to$EN \\
     & Whisper-Large-v3     + Chatterbox        & \texttt{whisper-large-v3}            + \texttt{ChatterboxMultilingualTTS}                                & Cascade                     & X$\to$EN \\
     & Seamless-Medium      + CosyVoice3        & \texttt{seamless-m4t-medium}         + \texttt{Fun-CosyVoice3-0.5B-2512}                                 & Cascade                     & X$\leftrightarrow$EN \\
     & Seamless-Medium      + Chatterbox        & \texttt{seamless-m4t-medium}         + \texttt{ChatterboxMultilingualTTS}                                & Cascade                     & X$\leftrightarrow$EN \\
     & Seamless-Large-v2    + CosyVoice3        & \texttt{seamless-m4t-v2-large}       + \texttt{Fun-CosyVoice3-0.5B-2512}                                 & Cascade                     & X$\leftrightarrow$EN \\
     & Seamless-Large-v2    + Chatterbox        & \texttt{seamless-m4t-v2-large}       + \texttt{ChatterboxMultilingualTTS}                                & Cascade                     & X$\leftrightarrow$EN \\
     & Gemma-4-E4B          + CosyVoice3        & \texttt{google/gemma-4-E4B}          + \texttt{Fun-CosyVoice3-0.5B-2512}                                 & Cascade                     & X$\leftrightarrow$EN \\
     & Gemma-4-E4B          + Chatterbox        & \texttt{google/gemma-4-E4B}          + \texttt{ChatterboxMultilingualTTS}                                & Cascade                     & X$\leftrightarrow$EN \\
     & Voxtral-Small        + CosyVoice3        & \texttt{Voxtral-Small-24B-2507}      + \texttt{Fun-CosyVoice3-0.5B-2512}                                 & Cascade                     & X$\leftrightarrow$EN \\
     & Voxtral-Small        + Chatterbox        & \texttt{Voxtral-Small-24B-2507}      + \texttt{ChatterboxMultilingualTTS}                                & Cascade                     & X$\leftrightarrow$EN \\
     & Qwen2.5-Omni (S2TT)  + CosyVoice3        & \texttt{Qwen2.5-Omni-7B}             + \texttt{Fun-CosyVoice3-0.5B-2512}                                 & Cascade                     & X$\leftrightarrow$EN \\
     & Qwen2.5-Omni (S2TT)  + Chatterbox        & \texttt{Qwen2.5-Omni-7B}             + \texttt{ChatterboxMultilingualTTS}                                & Cascade                     & X$\leftrightarrow$EN \\
     & Qwen3-Omni   (S2TT)  + CosyVoice3        & \texttt{Qwen3-Omni-30B-A3B-Instruct} + \texttt{Fun-CosyVoice3-0.5B-2512}                                 & Cascade                     & X$\leftrightarrow$EN \\
     & Qwen3-Omni   (S2TT)  + Chatterbox        & \texttt{Qwen3-Omni-30B-A3B-Instruct} + \texttt{ChatterboxMultilingualTTS}                                & Cascade                     & X$\leftrightarrow$EN \\
    \midrule
    \multirow{24}{*}{\rotatebox{90}{ASR + MT + TTS}}
     & Whisper-v2 + NLLB         + CosyVoice    & \texttt{whisper-large-v2}         + \texttt{nllb-200-3.3B}        + \texttt{Fun-CosyVoice3-0.5B-2512}    & Cascade                     & X$\leftrightarrow$EN \\
     & Whisper-v2 + NLLB         + Chatterbox   & \texttt{whisper-large-v2}         + \texttt{nllb-200-3.3B}        + \texttt{ChatterboxMultilingualTTS}   & Cascade                     & X$\leftrightarrow$EN \\
     & Whisper-v2 + Gemma-3-27B  + CosyVoice    & \texttt{whisper-large-v2}         + \texttt{gemma-3-27b-it}       + \texttt{Fun-CosyVoice3-0.5B-2512}    & Cascade                     & X$\leftrightarrow$EN \\
     & Whisper-v2 + Gemma-3-27B  + Chatterbox   & \texttt{whisper-large-v2}         + \texttt{gemma-3-27b-it}       + \texttt{ChatterboxMultilingualTTS}   & Cascade                     & X$\leftrightarrow$EN \\
     & Whisper-v2 + Qwen3-8B     + CosyVoice    & \texttt{whisper-large-v2}         + \texttt{Qwen3-8B}             + \texttt{Fun-CosyVoice3-0.5B-2512}    & Cascade                     & X$\leftrightarrow$EN \\
     & Whisper-v2 + Qwen3-8B     + Chatterbox   & \texttt{whisper-large-v2}         + \texttt{Qwen3-8B}             + \texttt{ChatterboxMultilingualTTS}   & Cascade                     & X$\leftrightarrow$EN \\
     & Whisper-v3 + NLLB         + CosyVoice    & \texttt{whisper-large-v3}         + \texttt{nllb-200-3.3B}        + \texttt{Fun-CosyVoice3-0.5B-2512}    & Cascade                     & X$\leftrightarrow$EN \\
     & Whisper-v3 + NLLB         + Chatterbox   & \texttt{whisper-large-v3}         + \texttt{nllb-200-3.3B}        + \texttt{ChatterboxMultilingualTTS}   & Cascade                     & X$\leftrightarrow$EN \\
     & Whisper-v3 + Gemma-3-27B  + CosyVoice    & \texttt{whisper-large-v3}         + \texttt{gemma-3-27b-it}       + \texttt{Fun-CosyVoice3-0.5B-2512}    & Cascade                     & X$\leftrightarrow$EN \\
     & Whisper-v3 + Gemma-3-27B  + Chatterbox   & \texttt{whisper-large-v3}         + \texttt{gemma-3-27b-it}       + \texttt{ChatterboxMultilingualTTS}   & Cascade                     & X$\leftrightarrow$EN \\
     & Whisper-v3 + Qwen3-8B     + CosyVoice    & \texttt{whisper-large-v3}         + \texttt{Qwen3-8B}             + \texttt{Fun-CosyVoice3-0.5B-2512}    & Cascade                     & X$\leftrightarrow$EN \\
     & Whisper-v3 + Qwen3-8B     + Chatterbox   & \texttt{whisper-large-v3}         + \texttt{Qwen3-8B}             + \texttt{ChatterboxMultilingualTTS}   & Cascade                     & X$\leftrightarrow$EN \\
     & Seamless-M  + NLLB         + CosyVoice   & \texttt{seamless-m4t-medium}      + \texttt{nllb-200-3.3B}        + \texttt{Fun-CosyVoice3-0.5B-2512}    & Cascade                     & X$\leftrightarrow$EN \\
     & Seamless-M  + NLLB         + Chatterbox  & \texttt{seamless-m4t-medium}      + \texttt{nllb-200-3.3B}        + \texttt{ChatterboxMultilingualTTS}   & Cascade                     & X$\leftrightarrow$EN \\
     & Seamless-M  + Gemma-3-27B  + CosyVoice   & \texttt{seamless-m4t-medium}      + \texttt{gemma-3-27b-it}       + \texttt{Fun-CosyVoice3-0.5B-2512}    & Cascade                     & X$\leftrightarrow$EN \\
     & Seamless-M  + Gemma-3-27B  + Chatterbox  & \texttt{seamless-m4t-medium}      + \texttt{gemma-3-27b-it}       + \texttt{ChatterboxMultilingualTTS}   & Cascade                     & X$\leftrightarrow$EN \\
     & Seamless-M  + Qwen3-8B     + CosyVoice   & \texttt{seamless-m4t-medium}      + \texttt{Qwen3-8B}             + \texttt{Fun-CosyVoice3-0.5B-2512}    & Cascade                     & X$\leftrightarrow$EN \\
     & Seamless-M  + Qwen3-8B     + Chatterbox  & \texttt{seamless-m4t-medium}      + \texttt{Qwen3-8B}             + \texttt{ChatterboxMultilingualTTS}   & Cascade                     & X$\leftrightarrow$EN \\
     & Seamless-Lv2 + NLLB         + CosyVoice  & \texttt{seamless-m4t-v2-large}    + \texttt{nllb-200-3.3B}        + \texttt{Fun-CosyVoice3-0.5B-2512}    & Cascade                     & X$\leftrightarrow$EN \\
     & Seamless-Lv2 + NLLB         + Chatterbox & \texttt{seamless-m4t-v2-large}    + \texttt{nllb-200-3.3B}        + \texttt{ChatterboxMultilingualTTS}   & Cascade                     & X$\leftrightarrow$EN \\
     & Seamless-Lv2 + Gemma-3-27B  + CosyVoice  & \texttt{seamless-m4t-v2-large}    + \texttt{gemma-3-27b-it}       + \texttt{Fun-CosyVoice3-0.5B-2512}    & Cascade                     & X$\leftrightarrow$EN \\
     & Seamless-Lv2 + Gemma-3-27B  + Chatterbox & \texttt{seamless-m4t-v2-large}    + \texttt{gemma-3-27b-it}       + \texttt{ChatterboxMultilingualTTS}   & Cascade                     & X$\leftrightarrow$EN \\
     & Seamless-Lv2 + Qwen3-8B     + CosyVoice  & \texttt{seamless-m4t-v2-large}    + \texttt{Qwen3-8B}             + \texttt{Fun-CosyVoice3-0.5B-2512}    & Cascade                     & X$\leftrightarrow$EN \\
     & Seamless-Lv2 + Qwen3-8B     + Chatterbox & \texttt{seamless-m4t-v2-large}    + \texttt{Qwen3-8B}             + \texttt{ChatterboxMultilingualTTS}   & Cascade                     & X$\leftrightarrow$EN \\
    \bottomrule
    \end{tabular}}
    \caption{Full list of systems benchmarked in COMPASS, organized by architectural type. \texttt{CoM} indicates Chain-of-Modality. Checkpoints are linked in the corresponding subsection of this appendix.}
    \label{tab:systems}
\end{table*}

\section{Benchmarked Systems: Detailed Description}
\label{app:systems}
This section provides a detailed description of all S2ST systems benchmarked in this work, grouped by architectural type, together with the exact checkpoints and inference settings used.
Table~\ref{tab:systems} summarizes the full set of systems and configurations.

\subsection{End-to-End Systems}
End-to-end S2ST models map source speech directly to target speech within a single neural architecture, without \textit{explicitly} producing intermediate text in the standard inference path.
We benchmark three families of end-to-end models.

\paragraph{SeamlessM4T.}
SeamlessM4T~\cite{barrault2023seamlessm4t, barrault2023seamless} is a multilingual and multitask foundation model that supports speech recognition, speech-to-text translation, text-to-text translation, and speech-to-speech translation.
We use the \texttt{Medium}\footnote{\href{https://huggingface.co/facebook/seamless-m4t-medium}{\texttt{huggingface.co/facebook/seamless-m4t-medium}}} (1.2B parameters) and \texttt{Large-v2}\footnote{\href{https://huggingface.co/facebook/seamless-m4t-v2-large}{\texttt{huggingface.co/facebook/seamless-m4t-v2-large}}} (2.3B parameters) checkpoints, both released by Meta AI.
Inference is run with the official \texttt{seamless\_communication} library,\footnote{\href{https://github.com/facebookresearch/seamless_communication}{\texttt{github.com/facebookresearch/seamless\_comm.}}} with default beam search and language tokens set explicitly for both source and target.
SeamlessM4T provides a vocoder that directly generates target audio from intermediate speech units, enabling fully end-to-end S2ST. We use the default speaker settings.

\paragraph{Qwen2.5-Omni.}
Qwen2.5-Omni~\cite{xu2025qwen25omnitechnicalreport} is a multimodal large language model that natively handles text, audio, image, and video inputs and produces both text and speech outputs.
We use the 7B checkpoint officially released by Alibaba.\footnote{\href{https://huggingface.co/Qwen/Qwen2.5-Omni-7B}{\texttt{huggingface.co/Qwen/Qwen2.5-Omni-7B}}}
We benchmark Qwen2.5-Omni in two inference modes:
(i) \texttt{Standard inference}, where a single prompt instructs the model to translate the input speech and produce the target speech directly;
(ii) \texttt{Chain-of-modality (CoM)}, where the same model is prompted to perform ASR, MT, and TTS sequentially, all within a single forward pass. This isolates each sub-task and tests whether explicit decomposition improves S2ST quality, while still using a single underlying model.

\paragraph{Qwen3-Omni.}
Qwen3-Omni~\cite{xu2025qwen3} is the successor to Qwen2.5-Omni, with improved multilingual coverage and audio generation quality.
We use the 30B released version (3B active parameters).\footnote{\href{https://huggingface.co/Qwen/Qwen3-Omni-30B-A3B-Instruct}{\texttt{huggingface.co/Qwen/Qwen3-Omni-30B-A3B-IT}}}
As with Qwen2.5-Omni, we benchmark both \texttt{standard} and \texttt{chain-of-modality} inference modes.

\subsection{Two-Stage Cascades: S2TT + TTS}

In a two-stage cascade, a speech-to-text translation model directly produces target text from source speech, and a TTS model synthesizes the target audio from this text.
This pipeline avoids the explicit ASR step but still relies on a separate TTS module.

\paragraph{S2TT components.} We employ specific task prompts and conditioning instructions for the model backbones, strictly following the configurations detailed in their respective official repositories:
\begin{itemize}[noitemsep,topsep=5pt,partopsep=0pt]
    \item \textbf{Whisper-Large-v2 / v3}~\cite{radford2023robust}: encoder-decoder model trained on 680k hours of multilingual audio.\footnote{\href{https://huggingface.co/openai/whisper-large-v2}{\texttt{huggingface.co/openai/whisper-large-v2}}}\footnote{\href{https://huggingface.co/openai/whisper-large-v3}{\texttt{huggingface.co/openai/whisper-large-v3}}} Whisper supports speech translation \textit{only into English}; it is therefore included in the X$\to$EN cascades only.
    \item \textbf{SeamlessM4T Medium / Large-v2}: same checkpoints as above, used in S2TT mode (no vocoder).
    \item \textbf{Gemma-4-E4B}~\cite{gemma4}: a lightweight multimodal model from Google DeepMind, natively supporting audio input. We use the \texttt{Instruction-Tuned} variant.\footnote{\href{https://huggingface.co/google/gemma-4-E4B-it}{\texttt{huggingface.co/google/gemma-4-E4B-it}}}
    \item \textbf{Voxtral-Small}~\cite{liu2025voxtral}: a 24B speech-language model from Mistral AI, mainly designed for speech recognition and translation.\footnote{\href{https://huggingface.co/mistralai/Voxtral-Small-24B-2507}{\texttt{huggingface.co/mistralai/Voxtral-Small-24B}}}
    \item \textbf{Qwen2.5-Omni / Qwen3-Omni}: same checkpoints as above, used in S2TT-only mode (audio output disabled).
\end{itemize}

\paragraph{TTS components.} Synthesis parameters, multi-speaker embeddings, and style conditioning tokens are controlled using the default generation configurations specified in the official model repositories:
\begin{itemize}[noitemsep,topsep=5pt,partopsep=0pt]
    \item \textbf{CosyVoice3}~\cite{cosyvoice}: a multilingual zero-shot TTS system with strong cross-lingual voice cloning capability.\footnote{\href{https://huggingface.co/FunAudioLLM/Fun-CosyVoice3-0.5B-2512}{\texttt{huggingface.co/FunAudioLLM/Fun-CosyVoice3-0.5B}}}
    \item \textbf{Chatterbox-TTS}~\cite{chatterbox}: an open-source emotion-controllable and speaker-conditioning TTS model.\footnote{\href{https://huggingface.co/ResembleAI/chatterbox}{\texttt{huggingface.co/ResembleAI/chatterbox}}} We use the English-only version for the generation of English targets, and the multilingual variant for other languages.
\end{itemize}
For both TTS systems, we use a fixed reference voice prompt taken from the source utterance to enable voice preservation, and we evaluate the resulting speaker similarity as one of the COMPASS metrics.

\subsection{Three-Stage Cascades: ASR + MT + TTS}
In the most explicit pipeline, source speech is first transcribed (ASR), the transcript is then translated (MT), and the translation is finally synthesized (TTS).
This decomposition is the historical baseline of S2ST and remains competitive thanks to strong individual components.

\paragraph{ASR components.} Transcription and alignment pipelines utilize language-conditioned decoding and inference arguments aligned with the optimal defaults outlined in the official implementations:
\begin{itemize}[noitemsep,topsep=5pt,partopsep=0pt]
    \item \textbf{Whisper-Large-v2 / v3}: same checkpoints as above, used for source-language transcription (multilingual ASR, not translation).
    \item \textbf{SeamlessM4T Medium / Large-v2}: same checkpoints as above, used in ASR mode.
\end{itemize}

\paragraph{MT components.} Textual translation modules utilize standard decoding hyperparameters specified by each system's official codebase:
\begin{itemize}[noitemsep,topsep=5pt,partopsep=0pt]
    \item \textbf{NLLB-200}~\cite{costa2022no}: a dedicated multilingual MT model from Meta. We use the 3.3B distilled variant.\footnote{\href{https://huggingface.co/facebook/nllb-200-3.3B}{\texttt{huggingface.co/facebook/nllb-200-3.3B}}}
    \item \textbf{Gemma-3-27B}~\cite{kamath2025gemma}: a multilingual instruction-tuned LLM from Google, prompted with the source-language transcript and a translation instruction. We use the \texttt{Instruction-Tuned} variant.\footnote{\href{https://huggingface.co/google/gemma-3-27b-it}{\texttt{huggingface.co/google/gemma-3-27b-it}}}
    \item \textbf{Qwen3-8B}~\cite{yang2025qwen3}: a general-purpose multilingual LLM from Alibaba, prompted to translate the ASR transcript.\footnote{\href{https://huggingface.co/Qwen/Qwen3-8B}{\texttt{huggingface.co/Qwen/Qwen3-8B}}}
\end{itemize}

\paragraph{TTS components.}
The same TTS models as in the 2-stage cascades: CosyVoice3 and Chatterbox.

\subsection{Inference Settings}
Across all systems, we use the following inference settings to ensure deterministic and comparable results:
\begin{itemize}[noitemsep,topsep=5pt,partopsep=0pt]
    \item Greedy decoding for all neural components, unless a model's official recipe explicitly recommends beam search (in which case we use the recommended beam size).
    \item Fixed random seeds (\texttt{seed=42}).
    \item Default sampling rates: 16 kHz for ASR/S2TT inputs and intermediate audio, with TTS outputs resampled to 16 kHz prior to metric computation.
    \item No system-specific post-processing (e.g., text normalization is applied uniformly, see Appendices~\ref{app:compass-toolkit} and~\ref{app:metrics}).
\end{itemize}

\subsection{Configuration Count}
The total number of model--language configurations is summarized below.
\begin{itemize}[noitemsep,topsep=5pt,partopsep=0pt]
    \item \textit{FLEURS X$\to$EN}: 46 systems $\times$ 10 language pairs $=$ 460 configurations.
    \item \textit{FLEURS EN$\to$X}: 42 systems $\times$ 10 language pairs $=$ 420 configurations.
    \item \textit{CVSS X$\to$EN}: 46 systems $\times$ 8 language pairs $=$ 368 configurations.
    \item \textit{Total}: $1,248$ configurations.
\end{itemize}

\section{The COMPASS Filtering Pipeline}
\label{app:filtering}
Fig.~\ref{fig:filtering} in the main document illustrates the full filtering pipeline. Below, we describe each step in detail.

\paragraph{Step 1: System-level aggregation.}
For each metric, we aggregate per-utterance scores into system-level scores by averaging across utterances and languages within each language group.

\paragraph{Step 2: Pairwise correlation.}
We compute the Spearman rank correlation $\rho$ between every pair of metrics across systems.
We decided to employ Spearman rather than Pearson because it is robust to scale differences and non-linear monotonic relationships.

\paragraph{Step 3: Hierarchical clustering.}
We convert the correlation matrix into a distance matrix $D = 1 - |\rho|$ and apply Ward-linkage hierarchical clustering.
Clusters are formed at a distance threshold of $0.15$, equivalent to grouping metrics with $|\rho| > 0.85$. Each resulting cluster contains metrics that carry essentially the same signal across systems.

\paragraph{Step 4: Multi-criteria scoring.}
For each metric, we compute four complementary scores.
\begin{itemize}[noitemsep,topsep=5pt,partopsep=0pt]
    \item \textit{Discriminability} ($D_m$): the coefficient of variation of the metric across systems, capturing how well the metric separates systems.
    \item \textit{Cross-language stability} ($C_m$): the mean Spearman $\rho$ of system rankings induced by the metric across pairs of languages, capturing whether the metric agrees on which system is better regardless of language.
    \item \textit{Independence} ($I_m$): $1 - \overline{|\rho_m|}$, where $\overline{|\rho_m|}$ is the mean absolute correlation of metric $m$ with all other metrics. Higher values indicate that the metric carries information not already captured elsewhere.
    \item \textit{Literature adoption} ($L_m$): a normalized usage score derived from a literature scan of S2ST and speech translation papers, preserving comparability with prior work even when a metric is partially redundant.
\end{itemize}
Each score is min-max normalized to $[0,1]$ and combined as described in Equation~\ref{eq:score}. To ensure robustness, we conducted a sensitivity analysis by perturbing each weight by $\pm 20\%$; the resulting top metric rankings remained highly stable (Spearman's $\rho > 0.80$), justifying the unweighted formulation.

\paragraph{Step 5: Cluster representative selection.}
Within each redundancy cluster, we select the metrics with the highest recommendation score as the cluster representatives. The number of representatives scales with cluster size: dimensions with up to four metrics keep one representative, those with five to ten keep two, and larger dimensions keep three.

\paragraph{Step 6: Iterative group refinement.}
The candidate set from Step 5 is internally non-redundant within each a priori dimension, but representatives from \textit{different} dimensions may still carry overlapping signal.
For example, speaker-similarity, prosody, and articulation representatives may all move together across systems, even though they were defined as separate dimensions \textit{a priori}.
We therefore refine the grouping of representatives iteratively until the resulting groups are both internally coherent and mutually distinct:
\begin{itemize}[noitemsep,topsep=5pt,partopsep=0pt]
    \item \textit{Silhouette score}: computed in the space induced by system-level score vectors.
    A low silhouette ($< 0.25$) indicates that at least two groups carry redundant signal.
    \item \textit{Cross-group correlation} ($\rho_\text{cross}$): the mean absolute Spearman correlation between representatives from different groups. A high value ($> 0.70$) indicates two groups are statistically 
    interchangeable.
\end{itemize}
While silhouette $< 0.25$ or $\rho_\text{cross} > 0.70$, we merge the two most correlated groups, re-run Steps 2-5 within the merged group, and re-evaluate. The procedure stops when silhouette $\geq 0.25$ and all cross-group correlations are $\leq 0.70$, yielding a final compact set of non-redundant representatives organized into well-separated empirical dimensions.

\section{Detailed Benchmark Results}
\label{app:results}

\begin{table*}[!t]
    \centering
    \setlength{\tabcolsep}{4pt}
    \renewcommand{\arraystretch}{0.96}
    \resizebox{\textwidth}{!}{%
    \begin{tabular}{c l r r r r r r r r r r}
        \toprule
        \textbf{Data} 
            & \textbf{Statistic} 
            & \textbf{ar} 
            & \textbf{de} 
            & \textbf{es} 
            & \textbf{fr} 
            & \textbf{hi} 
            & \textbf{it} 
            & \textbf{ja} 
            & \textbf{ko} 
            & \textbf{pt} 
            & \textbf{zh} \\
        \midrule
        \multirow{5}{*}{\rotatebox{90}{\small FLEURS X$\rightarrow$EN}}
            & \textbf{\#Samples}         & 416    & 746    & 879    & 662    & 399    & 743    & 620    & 370    & 871    & 906    \\
            & \textbf{Avg Src Len}       & 108.78 & 143.61 & 148.73 & 151.51 & 120.28 & 140.34 &  51.73 &  60.86 & 134.24 &  40.91 \\
            & \textbf{Avg Tgt Len}       & 124.90 & 124.68 & 126.13 & 126.88 & 123.72 & 120.26 & 122.18 & 124.72 & 125.05 & 128.40 \\
            & \textbf{Avg Dur (s)}       &  10.60 &  11.22 &  11.90 &  10.10 &  11.03 &  13.05 &  12.62 &  12.26 &  12.04 &  11.26 \\
            & \textbf{Tot Dur (h)}       &   1.22 &   2.33 &   2.90 &   1.86 &   1.22 &   2.69 &   2.17 &   1.26 &   2.91 &   2.83 \\
        \midrule \midrule
        \multirow{5}{*}{\rotatebox{90}{\small FLEURS EN$\rightarrow$X}}
            & \textbf{\#Samples}     &  511   &  611   &  622   &  600   &  463   &  590   &  566   &  485   &  618   &  633   \\
            & \textbf{Avg Src Len}   & 126.52 & 127.59 & 128.33 & 128.44 & 123.86 & 124.92 & 125.14 & 126.16 & 127.23 & 129.81 \\
            & \textbf{Avg Tgt Len}   & 110.56 & 147.55 & 151.58 & 153.47 & 121.40 & 147.01 &  53.39 &  62.38 & 136.77 &  41.73 \\
            & \textbf{Avg Dur (s)}   &   9.32 &   9.50 &   9.56 &   9.58 &   9.30 &   9.38 &   9.41 &   9.47 &   9.51 &   9.64 \\
            & \textbf{Tot Dur (h)}   &   1.32 &   1.61 &   1.65 &   1.60 &   1.20 &   1.54 &   1.48 &   1.28 &   1.63 &   1.70 \\
        \midrule \midrule
        \multirow{5}{*}{\rotatebox{90}{\small CVSS X$\rightarrow$EN}}
            & \textbf{\#Samples}         & 1000   & 1000   & 1000   & 1000   & -      & 1000   &  684   & -      & 1000   & 1000  \\
            & \textbf{Avg Src Len}       & 23.35  & 63.02  & 61.68  & 59.57  & -      & 64.05  & 19.17  & -      & 46.45  & 19.19 \\
            & \textbf{Avg Tgt Len}       & 30.34  & 58.95  & 61.00  & 55.78  & -      & 61.29  & 45.19  & -      & 43.54  & 70.08 \\
            & \textbf{Avg Dur (s)}       &  3.90  &  5.66  &  6.15  &  5.54  & -      &  6.12  &  4.62  & -      &  4.65  &  5.99 \\
            & \textbf{Tot Dur (h)}       &  1.08  &  1.57  &  1.71  &  1.54  & -      &  1.70  &  0.88  & -      &  1.29  &  1.67 \\
        \bottomrule
    \end{tabular}}
    \caption{Per-language data statistics: number of samples, average source/target text length (characters), average audio duration, and total audio duration. Note that, for CVSS, we only use the first 1,000 utterances per pair in our experiments.}
    \label{tab:data_stats}
\end{table*}
\subsection{Dataset Statistics}
\label{app:datasets_statistics}
Table~\ref{tab:data_stats} reports per-language statistics for FLEURS (both directions) and CVSS (X$\to$EN). FLEURS provides between 370 (\texttt{ko}) and 906 (\texttt{zh}) samples per language in X$\to$EN, and 463-633 in EN$\to$X, with consistent average durations of 9-13s per clip. Total audio ranges from 1.2h to 2.9h per language in X$\to$EN and 1.2h to 1.7h in EN$\to$X. CVSS contributes shorter clips (3.9-6.2s) capped at 1,000 samples per pair, totalling 0.9-1.7h per language; \texttt{hi} and \texttt{ko} are excluded as they are absent in CVSS. Source-text length varies across scripts, with CJK languages (\texttt{ja}, \texttt{zh}, \texttt{ko}) showing the lowest character counts due to denser orthography.

\subsection{Compact Metric Subsets}
\label{app:compact_metrics}
\begin{figure}[t]
    \centering
    \includegraphics[width=0.9\linewidth]{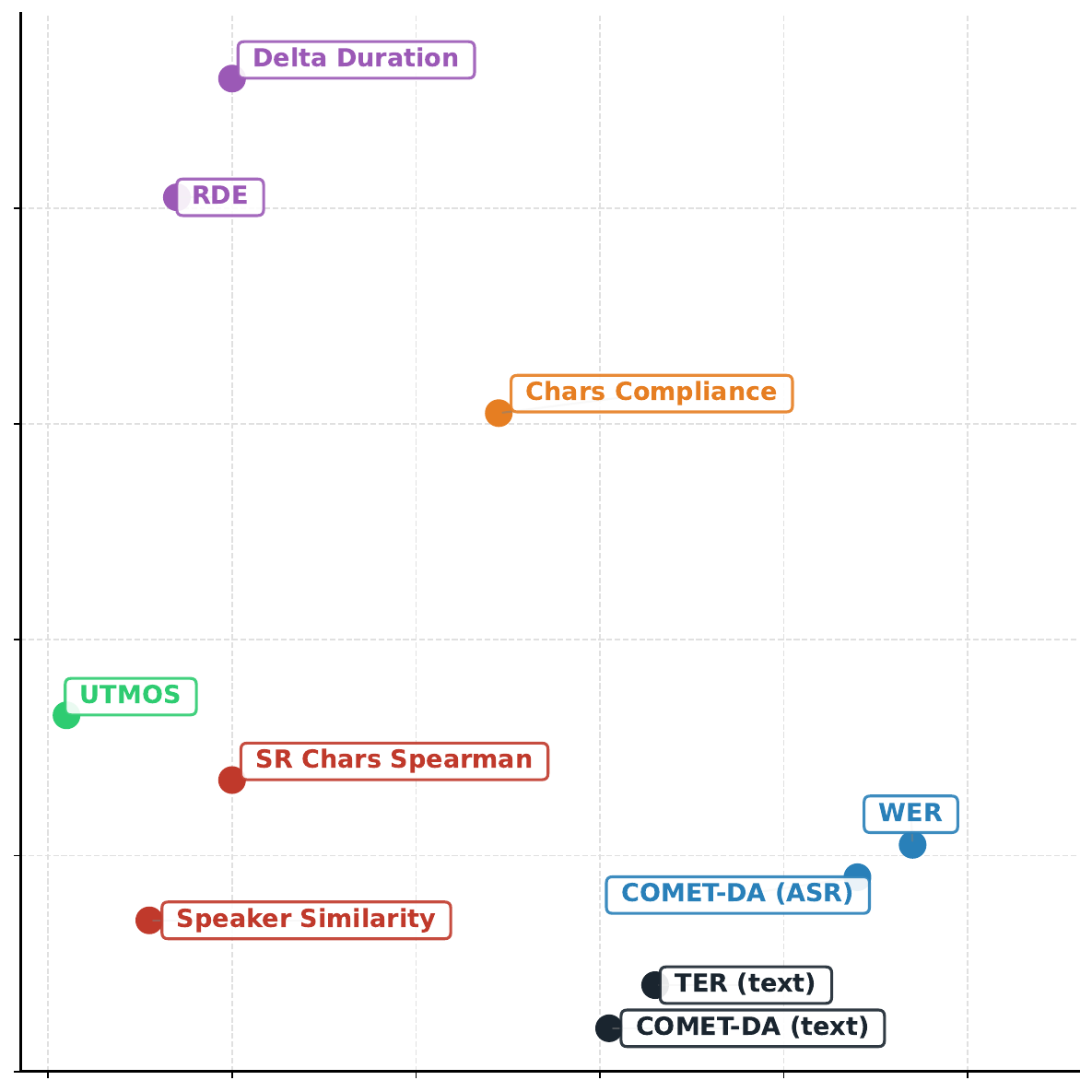} \quad
    \includegraphics[width=0.9\linewidth]{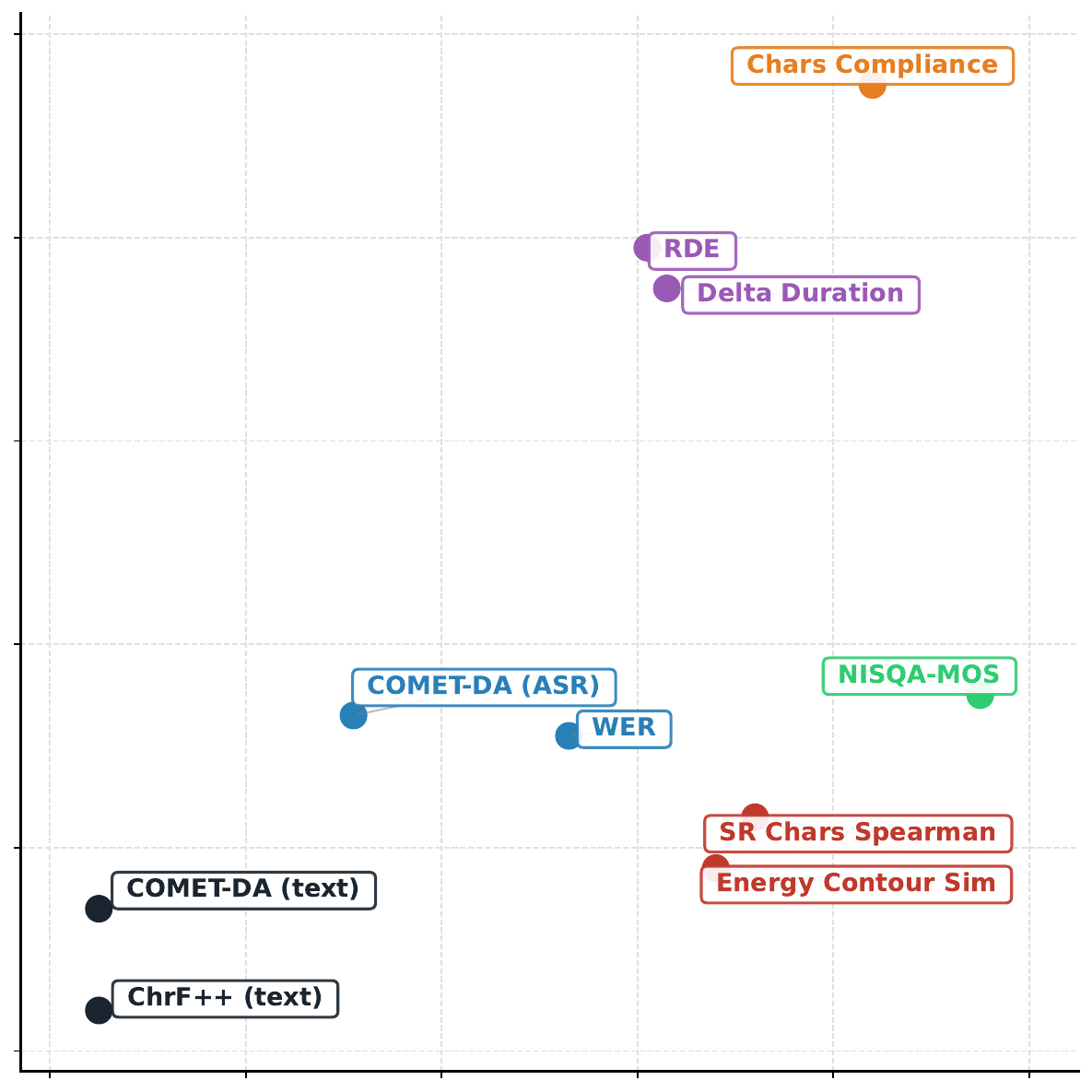}
    \caption{Two-dimensional $t$-SNE visualization of the full S2ST metric space for X$\to$EN (up) and EN$\to$X (bottom) translation directions. Metrics are colored according to their primary evaluation dimension. The clear spatial separation into disjoint, well-defined clusters highlights that the COMPASS filtering pipeline successfully identifies distinct orthogonal dimensions while minimizing intra-dimension structural redundancy.}
    \label{fig:subset-metrics-en-x-en}
    \vspace{-3mm}
\end{figure}

Table~\ref{tab:compact_metrics} in the main text reports the final compact metric subsets produced by the COMPASS filtering pipeline for the X$\to$EN and EN$\to$X translation directions. To visually validate the separation of these evaluation dimensions, we present a $t$-SNE clustering analysis of the metrics in Figure~\ref{fig:subset-metrics-en-x-en} (up for X$\to$EN, bottom for EN$\to$X). The projections show that the metrics organize into distinct, spatially isolated clusters corresponding to their underlying evaluation dimensions (e.g., translation quality for text in black and in blue for ASR, acoustic naturalness in green, or speech timing alignment in violet), confirming that our pipeline successfully filters out localized redundancy while retaining comprehensive domain coverage.

\begin{figure*}
    \centering
    \includegraphics[width=\linewidth]{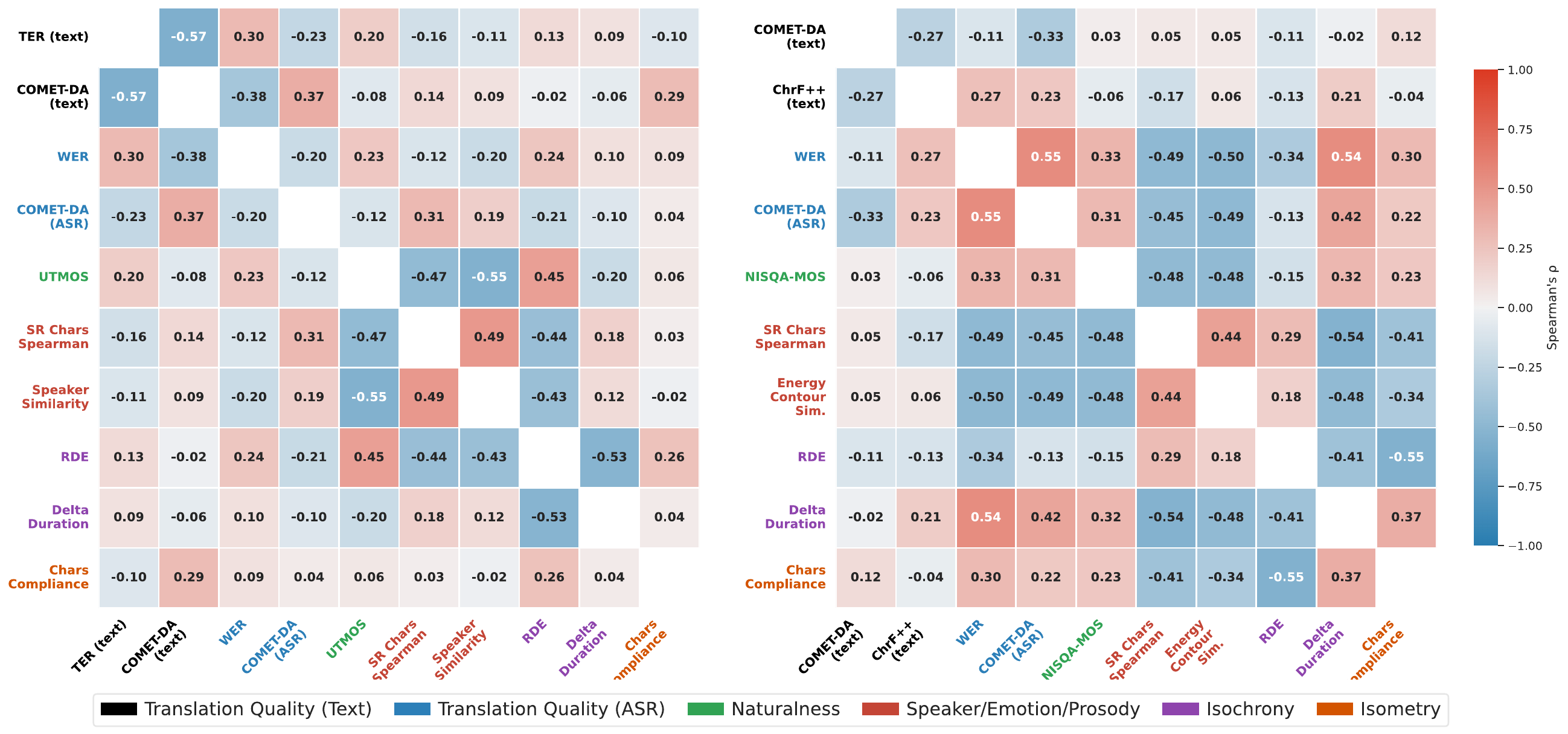}
    \caption{Pairwise Spearman rank correlation ($\rho$) matrices between metrics selected for the final compact subsets across X$\to$EN (left) and EN$\to$X (right) directions. The matrix demonstrates low cross-dimensional correlations. This confirms that the selected metrics deliver independent, complementary signals regarding system performance.}    
    \label{fig:spearman-rho-subsets}%
\end{figure*}

Furthermore, Fig.~\ref{fig:spearman-rho-subsets} displays the pairwise Spearman rank correlation matrices between the selected metrics for both translation directions (left for X$\to$EN, right for EN$\to$X). These matrices demonstrate low cross-dimensional correlation among the final selected subsets, confirming that the remaining components capture complementary information rather than redundant system characteristics.

\begin{table}[t]
    \centering
    \setlength{\tabcolsep}{4pt}
    \resizebox{0.96\columnwidth}{!}{%
    \begin{tabular}{l c c}
        \toprule
        \textbf{Pair (vs. \texttt{spk\_sim})} & \textbf{EN$\to$JA} & \textbf{JA$\to$EN} \\
        \midrule
        \texttt{AutoPCP}                   & +0.80 & +0.71 \\
        Emotion embedding sim.             & +0.90 & +0.89 \\
        F0 contour sim.                    & -0.70 & +0.94 \\
        Energy contour sim.                & -0.70 & +0.66 \\
        \midrule
        Mean $|\rho|$ (embedding-based)    & 0.70 & 0.80 \\
        \bottomrule
    \end{tabular}}
    \caption{Pairwise Spearman $\rho$ between speaker similarity and prosody/emotion metrics on MELD-ST.}
    \label{tab:meldst_corr}
\end{table}
\subsection{Read-Speech vs. Spontaneous Decoupling of Speaker and Prosody}
\label{app:meldst_decoupling}
To test whether the empirical merge of \textit{Speaker Consistency} and \textit{Prosody \& Emotion} (Sec.~\ref{sec:results}, RQ1) is corpus-dependent, we recompute pairwise Spearman $\rho$ correlations between the cluster's representative metrics on the spontaneous, emotionally rich MELD-ST corpus across our benchmarked systems (Q3O, SeamL, S2TT+TTS, ASR+MT+TTS, and ground-truth references, see Section~\ref{app:human_eval}).\footnote{We only employ the subset that is used for the human evaluation, containing around 133 samples per direction, later merged in 25 unique videos.}

\paragraph{Setup.}
We aggregate utterance-level scores to the system level for the speaker metric (\texttt{Speaker Similarity}) and three prosody/emotion metrics: \texttt{AutoPCP}, emotion-embedding similarity, and F0/energy contour similarity. We then compute pairwise Spearman correlations across systems for each translation direction (EN$\to$JA and JA$\to$EN).

\paragraph{Results.}
Table~\ref{tab:meldst_corr} reports the per-pair correlations and aggregate mean $|\rho|$ values. On read-speech benchmarks (FLEURS, CVSS), the speaker-prosody cluster exhibits a mean $|\rho| > 0.85$, the threshold used by our filtering pipeline (Sec.~\ref{sec:filtering}) to merge correlated metrics. On MELD-ST, the average drops to $|\rho| \approx$ 0.75 (mean of EN$\to$JA: 0.70, JA$\to$EN: 0.80). Prosodic-contour metrics (F0, energy) decouple more sharply while embedding-based metrics (AutoPCP, emotion embedding) remain more strongly tied to speaker similarity.

\paragraph{Interpretation.}
The partial decoupling on spontaneous speech (mean $|\rho|$ moves from $>0.85$ on read corpora down to $0.70$-$0.80$ on MELD-ST) supports the view that the six-dimensional empirical collapse is partly driven by the limited prosodic variability of read-speech benchmarks. However, the residual link remains strong, indicating that contemporary S2ST systems also exhibit genuinely correlated speaker and prosody behavior, likely because voice-cloning components jointly transfer timbre and prosodic style. We therefore retain the eight-dimensional conceptual taxonomy and recommend re-estimating the empirical grouping as new architectures and more expressive corpora become available.

\paragraph{Caveats.}
The analysis uses N=5 systems per direction, so individual $\rho$ values are noisy. Our interpretation rests on the \textit{direction} and \textit{magnitude of the change} between read- and spontaneous-speech corpora, not on individual significance levels. More details on the human evaluation on~\ref{app:human_eval}.

\subsection{Cross-Language Stability of Selected Metrics}
\label{app:stability}
Table~\ref{tab:stability} reports the cross-language stability score $C_m$ for each selected representative metric within the COMPASS evaluation suite. As introduced in Section~\ref{sec:filtering}, a score of $C_m = 1.0$ indicates that the metric induces an identical relative ranking of target systems across every single language group tested, while $C_m = 0$ denotes completely uncorrelated system rankings between different language cohorts. 

\begin{table}[!ht]
    \centering
    \resizebox{\columnwidth}{!}{%
    \begin{tabular}{cllc}
    \toprule
    \textbf{Dir.} 
        & \textbf{Metric} 
        & \textbf{Dimension} 
        & \textbf{Score} \\
    \midrule
    \multirow{10}{*}{\rotatebox{90}{X$\rightarrow$EN}}
        & \texttt{COMET-DA}                     & TQ (ASR)        & 0.85 \\
        & \texttt{WER}                          & TQ (ASR)        & 0.89 \\
        & \texttt{COMET-DA}                     & TQ (Text)       & 0.85 \\
        & \texttt{TER}                          & TQ (Text)       & 0.72 \\
        & \texttt{UTMOS}                        & Naturalness     & 0.82 \\
        & \texttt{SR Chars $\rho$}              & Spk \& Prosody  & 0.79 \\
        & \texttt{Speaker Sim.}                 & Spk \& Prosody  & 0.91 \\
        & \texttt{RDE}                          & Isochrony       & 0.69 \\
        & \texttt{$\Delta$ Duration}            & Isochrony       & 0.71 \\
        & \texttt{Chars Compliance}             & Isometry        & 0.74 \\
    \midrule 
    \multirow{10}{*}{\rotatebox{90}{EN$\rightarrow$X}}
        & \texttt{COMET-DA}                     & TQ (ASR)        & 0.83 \\
        & \texttt{WER}                          & TQ (ASR)        & 0.85 \\
        & \texttt{COMET-DA}                     & TQ (Text)       & 0.83 \\
        & \texttt{ChrF++}                       & TQ (Text)       & 0.75 \\
        & \texttt{NISQA-MOS}                    & Naturalness     & 0.74 \\
        & \texttt{SR Chars $\rho$}              & Spk \& Prosody  & 0.76 \\
        & \texttt{Energy Cont. Sim.}            & Spk \& Prosody  & 0.85 \\
        & \texttt{RDE}                          & Isochrony       & 0.72 \\
        & \texttt{$\Delta$ Duration}            & Isochrony       & 0.73 \\
        & \texttt{Chars Compliance}             & Isometry        & 0.70 \\
    \bottomrule
    \end{tabular}}
    \caption{Cross-language stability ($C_m$) of selected COMPASS representative metrics per direction. Higher is better.}
    \label{tab:stability}
    \vspace{-3mm}
\end{table}

\begin{table*}[!ht]
    \centering
    \resizebox{\linewidth}{!}{%
    \begin{tabular}{cllccccccc}
    \toprule
    \multicolumn{2}{l}{\textbf{Data}}
        & \textbf{System} 
        & \textbf{COMET-DA (text)} 
        & \textbf{COMET-DA (ASR)} 
        & \textbf{Word Acc.} 
        & \textbf{UTMOS} 
        & \textbf{Spk Sim} 
        & \textbf{RDE} 
        & \textbf{Char Compl. (\%)} \\
    \midrule
    \multirow{4}{*}{\rotatebox{90}{\footnotesize FLEURS}}
    & \multirow{4}{*}{\rotatebox{90}{\footnotesize X$\rightarrow$EN}}
        & S2TT+TTS        
        & 0.861 
        & \textbf{0.878} 
        & \textbf{0.960} 
        & 3.071 
        & \textbf{0.915} 
        & -0.361 
        & 18.920 \\
    && ASR+MT+TTS   
        & 0.852 
        & 0.840         
        & 0.946 
        & 3.332 
        & 0.805          
        & \textbf{-0.243} 
        & 24.490 \\
    && SeamL                
        & 0.833 
        & 0.847          
        & 0.977 
        & 3.060 
        & 0.663          
        & -0.404 
        & 24.812 \\
    && Q3O                
        & \textbf{0.880} 
        & 0.863          
        & 0.942
        & \textbf{3.591} 
        & 0.606          
        & -0.334 
        & \textbf{24.833} \\
        
    \midrule\midrule
    
    \multicolumn{2}{l}{\textbf{Data}}
        &\textbf{System} 
        & \textbf{Text ChrF++} 
        & \textbf{COMET-DA (ASR)} 
        & \textbf{Word Acc.} 
        & \textbf{NISQA-MOS} 
        & \textbf{SR Char Sp.} 
        & \textbf{RDE} 
        & \textbf{Char Compl. (\%)} \\
    \midrule
    \multirow{4}{*}{\rotatebox{90}{\footnotesize FLEURS}}
    & \multirow{4}{*}{\rotatebox{90}{\footnotesize EN$\rightarrow$X}}
    & S2TT+TTS        
        & 54.062
        & \textbf{0.804} 
        & \textbf{0.781} 
        & 2.410
        & \textbf{0.498} 
        & \textbf{-0.006} 
        & 20.392 \\
    && ASR+MT+TTS   
        & \textbf{57.342} 
        & 0.716 
        & 0.650
        & 2.661 
        & 0.165 
        & -0.024 
        & 19.463 \\
    && SeamL                 
        & 49.931 
        & 0.798 
        & 0.756 
        & 3.654 
        & 0.169 
        & -0.147 
        & \textbf{21.040} \\
    && Q3O               
        & 51.321 
        & 0.777 
        & 0.746 
        & \textbf{4.321} 
        & 0.166 
        & -0.091 
        & 18.692 \\

    \midrule\midrule

    \multicolumn{2}{l}{\textbf{Data}}
        & \textbf{System} 
        & \textbf{COMET-DA (text)} 
        & \textbf{COMET-DA (ASR)} 
        & \textbf{Word Acc.} 
        & \textbf{UTMOS} 
        & \textbf{Spk Sim} 
        & \textbf{RDE} 
        & \textbf{Char Compl. (\%)} \\
    \midrule
    \multirow{4}{*}{\rotatebox{90}{\footnotesize CVSS}}
    & \multirow{4}{*}{\rotatebox{90}{\footnotesize X$\rightarrow$EN}}
    & S2TT+TTS        
        & 0.809 
        & 0.784          
        & 0.933 
        & 2.882          
        & \textbf{0.889} 
        & -0.391          
        & 28.881 \\
    && ASR+MT+TTS   
        & 0.646 
        & 0.641          
        & \textbf{0.966}          
        & 2.890          
        & 0.779          
        & \textbf{-0.327} 
        & 30.603 \\
    && SeamL                 
        & 0.814 
        & 0.803          
        & 0.965          
        & 3.131          
        & 0.524          
        & -0.437          
        & \textbf{33.292} \\
    && Q3O               
        & \textbf{0.838} 
        & \textbf{0.839} 
        & 0.941          
        & \textbf{3.492} 
        & 0.608          
        & -0.421          
        & 31.101 \\
    \bottomrule
    
    \end{tabular}}
    \caption{Average system-level scores on FLEURS X$\to$EN (top), EN$\to$X (middle), and CVSS X$\to$EN (bottom), across the target languages. Best per column in \textbf{bold}.}
    \label{tab:fleurs_covost_full}
    \vspace{-2mm}
\end{table*}

By analyzing the intersection of system rankings across geographically and structurally diverse language groupings (such as \textit{Romance} and \textit{CJK}), the $C_m$ score allows us to identify metrics that remain reliable globally, rather than those whose discriminability depends artificially on a specific language family. 
The empirical findings reveal several key trends regarding metric robustness.
First, ASR-grounded and neural text evaluation metrics show exceptionally high stability. Embedding-driven metrics like \texttt{COMET-DA} score consistently high ($0.83 \le C_m \le 0.85$) regardless of whether they are applied to raw text references or speech transcripts. \texttt{WER} exhibits strong global system tracking properties ($0.85$ for EN$\to$X and $0.89$ for X$\to$EN), highlighting its utility as a cross-lingual proxy for transcription accuracy.
Second, modeling target speaker traits displays high stability in the X$\to$EN direction, where \texttt{Speaker Similarity} achieves the highest overall consistency ($C_m = 0.91$). In the EN$\to$X direction, tracking dynamic energy curves via \texttt{Energy Contour Similarity} provides an equally dependable signal ($C_m = 0.85$), showing that speaker modeling metrics retain strong rank-preservation properties even when adapting to highly divergent target acoustic environments.
Finally, structural constraint metrics such as Relative Duration Error (\texttt{RDE}), absolute duration difference (\texttt{$\Delta$ Duration}), and character length compliance exhibit slightly lower but highly stable scores ranging from $0.69$ to $0.74$. Because these metrics are tied fundamentally to length and timing matching, their lower baseline relative to neural metrics stems from cross-lingual differences in structural density; however, their consistent performance across both directions confirms their global validity as independent evaluation tracks.

\subsection{Per-System Results}
\label{app:rq3_full}
This appendix provides a comprehensive analysis of the system-level evaluations introduced in Section~\ref{sec:results} (RQ3). We benchmark four representative systems spanning distinct architectural paradigms: \texttt{S2TT+TTS} (a two-stage cascade, Voxtral + ChatterboxTTS), \texttt{ASR+MT+TTS} (a three-stage cascade, Whisper-Lv3 + Gemma3 + Cosyvoice3), \texttt{SeamL} (an end-to-end model, Seamless-Lv2), and \texttt{Q3O} (a SpeechLLM, Qwen3-Omni). The aggregate findings across all ten target languages for FLEURS $X\to\text{EN}$ and $\text{EN}\to X$, alongside the eight target languages for CVSS $X\to\text{EN}$, are detailed in Table~\ref{tab:fleurs_covost_full}.

\paragraph{Translation Quality (Text and ASR).}
System rankings for translation quality show a strong sensitivity. On FLEURS $X\to\text{EN}$ (Table~\ref{tab:fleurs_covost_full} (top)), the end-to-end SpeechLLM model \texttt{Q3O} achieves the highest semantic text quality with a \texttt{COMET-DA (text)} score of 0.880, closely followed by the cascaded configurations \texttt{S2TT+TTS} (0.861) and \texttt{ASR+MT+TTS} (0.852). This changes on the CVSS $X\to\text{EN}$ set (Table~\ref{tab:fleurs_covost_full} (bottom)), where \texttt{Q3O} retains its lead (0.838) but the three-stage cascade \texttt{ASR+MT+TTS} collapses to 0.646.
This performance drop is primarily caused by decoding loops during text generation on specific Romance languages, where the model repeatedly outputs the same phrases (averaging a text COMET-DA of $\approx 0.39$ for Spanish, French, and Italian). This effectively exposes the vulnerability of cascaded architectures to error propagation. 

Conversely, when shifting to the $\text{EN}\to X$ direction on FLEURS (Table~\ref{tab:fleurs_covost_full} (middle)), the \texttt{Text ChrF++} metric ranks \texttt{ASR+MT+TTS} highest at 57.34, followed by \texttt{Q3O} (51.32). Yet, looking at the \texttt{COMET-DA (ASR)} metric, the ranking flips: the cascade \texttt{S2TT+TTS} takes the lead at 0.804, while \texttt{ASR+MT+TTS} drops to 0.716. This clear divergence underscores why analyzing text-level or ASR-level metrics in isolation can yield incomplete system profiles.

\begin{figure*}
    \centering
    \includegraphics[width=\linewidth]{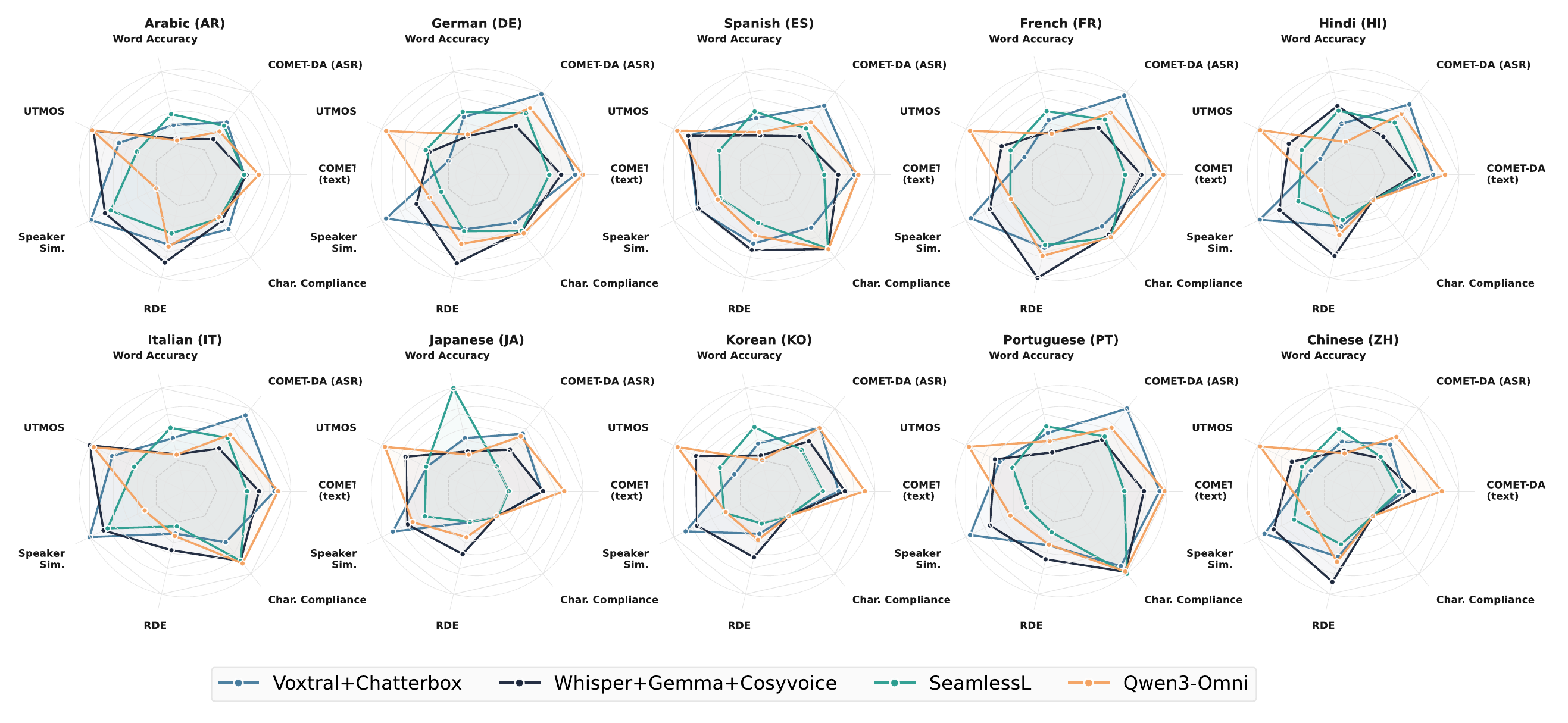}
    \caption{Per-language system performance profiles across the final filtered COMPASS dimensions on the FLEURS X$\rightarrow$EN evaluation set. Each radar plot displays model behavior across orthogonal axes corresponding to (a subset of) the representative metrics (including \texttt{COMET-DA}, 1-\texttt{WER} (\texttt{Word Accuracy}), \texttt{Speaker Sim.}, \texttt{UTMOS}, \texttt{RDE}, and \texttt{Chars Compliance}). This fine-grained layout highlights how structural system behaviors remain consistent across diverse source typologies, while isolating specific cross-lingual challenges such as character length compliance and speaker identity preservation across both close and distant language families.}    
    \label{fig:fleurs_x_en_radar_per_language}%
    \vspace{-3mm}
\end{figure*}

\paragraph{Audio Naturalness.}
The results show a systematic advantage for native end-to-end audio architectures when synthesizing target speech. For $\text{EN}\to X$ outputs on FLEURS, the end-to-end systems outperform their cascaded counterparts by a wide margin, with \texttt{Q3O} securing a \texttt{NISQA-MOS} of 4.32 and \texttt{SeamL} achieving 3.65. In comparison, the cascaded systems \texttt{ASR+MT+TTS} and \texttt{S2TT+TTS} trail at 2.66 and 2.41, respectively. A similar trend occurs in $X\to\text{EN}$ translation directions monitored via \texttt{UTMOS}. On FLEURS, \texttt{Q3O} leads with 3.59, followed by \texttt{ASR+MT+TTS} (3.33) and \texttt{S2TT+TTS} (3.07). On CVSS, \texttt{Q3O} still leads at 3.49, outperforming \texttt{SeamL} (3.13) and both cascades, which cluster around 2.88 and 2.89.

\paragraph{Speaker Identity and Prosodic Preservation.}
The architectural advantages invert when evaluating voice preservation, where cascaded systems display a decisive edge. For $X\to\text{EN}$ tasks, the voice-cloning capabilities embedded in dedicated TTS modules allow \texttt{S2TT+TTS} to dominate speaker similarity scores on both FLEURS (0.92) and CVSS (0.89). The three-stage cascade \texttt{ASR+MT+TTS} follows at 0.81 and 0.78. Meanwhile, the end-to-end models exhibit poor target speaker preservation, with \texttt{SeamL} scoring 0.66 / 0.52 and \texttt{Q3O} scoring 0.61 / 0.61. This disparity stems from a core design difference: end-to-end systems generally sample from a restricted internal pool of reference voices, making it difficult to capture individual speaker features dynamically. For $\text{EN}\to X$ tasks, tracking prosodic variation via \texttt{SR Char Sp.} confirms this trend; \texttt{S2TT+TTS} scores 0.498, while all other systems stall between 0.165 and 0.169, failing to mirror source speaking rate dynamics.

\paragraph{Isochrony (Temporal Alignment).}
On $X\to\text{EN}$, all evaluated systems show a systematic trend toward audio compression, as evidenced by uniformly negative Relative Duration Error (\texttt{RDE}) values. On FLEURS, \texttt{ASR+MT+TTS} exhibits the tightest temporal compliance with an \texttt{RDE} of $-0.243$, while \texttt{SeamL} shows the largest temporal discrepancy at $-0.404$. This relative ordering is maintained on the CVSS dataset. When translating from $\text{EN}\to X$, temporal errors narrow globally, and the cascades show near-perfect temporal matching. Specifically, \texttt{S2TT+TTS} reaches an exceptional alignment score of $-0.006$, followed by \texttt{ASR+MT+TTS} at $-0.024$, whereas \texttt{SeamL} lags at $-0.147$. These outcomes demonstrate that modular pipelines benefit heavily from explicit duration constraints in their generation headers, whereas end-to-end systems sacrifice structural timing accuracy to optimize global synthesis.

\begin{figure*}
    \centering
    \includegraphics[width=\linewidth]{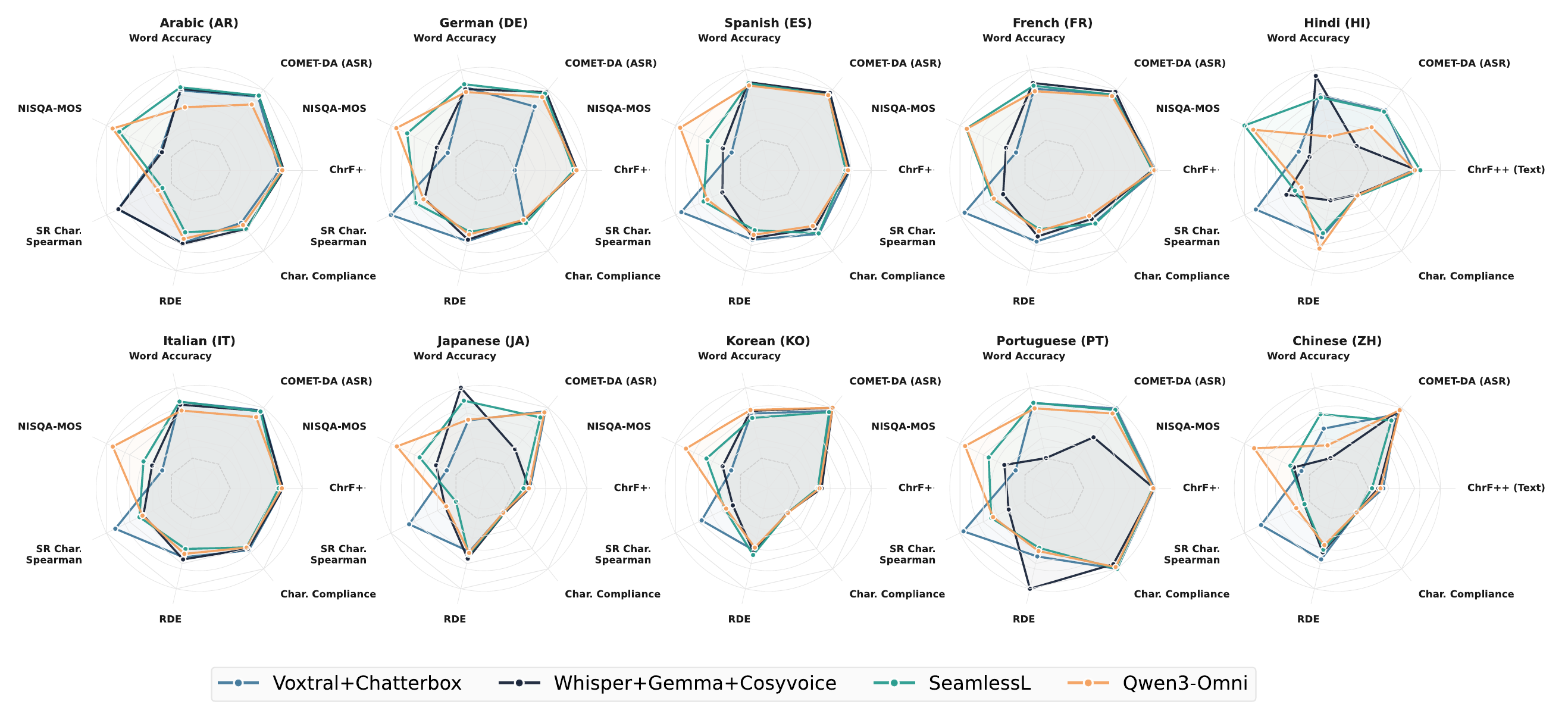}
    \caption{Per-language results for EN$\rightarrow$X translation on the FLEURS dataset. These plots show how different systems perform when generating output in diverse target languages, including the CJK (Chinese, Japanese, Korean) and Romance families. The shape of the plots confirms that our selected metrics, as \texttt{NISQA-MOS}, \texttt{ChrF++}, and \texttt{Energy Cont. Sim.}, work reliably across very different languages, keeping a clear distinction between the best and worst performing systems.}   \label{fig:fleurs_en_x_radar_per_language}%
    \vspace{-3mm}
\end{figure*}

\paragraph{Isometry (Length Compliance).}
Character length compliance (\texttt{Char Compl. (\%)}) remains consistently low across all datasets and models, spanning a tight interval from 18.69\% to 33.29\%. This baseline reveals that none of these core architectures is explicitly optimized for strict length matching. On FLEURS $X\to\text{EN}$, the end-to-end models and the text-cascaded pipeline perform within a tight window (24.49\%-24.83\%), whereas \texttt{S2TT+TTS} drops to 18.92\%. On FLEURS $\text{EN}\to X$, performance normalizes between 18.69\% and 21.04\%, led by \texttt{SeamL}. Turning to CVSS $X\to\text{EN}$, compliance increases slightly across the board, with \texttt{SeamL} leading at 33.29\% and \texttt{S2TT+TTS} following at 28.88\%. The narrow variation within each dataset suggests that isometric compliance is driven primarily by data-density attributes rather than by underlying model mechanics, leaving explicit length-control mechanisms as a key challenge for speech-to-speech translation research.

\paragraph{Summary of Findings.}
The multi-dimensional evaluation reveals a clear architectural trade-off: cascaded pipelines provide superior speaker preservation and precise structural duration control, but at the expense of output speech naturalness. In contrast, end-to-end systems deliver highly natural speech patterns but fail to reliably mirror source voice identity or timing boundaries. Translation quality, instead, remains competitive between both paradigms but shifts significantly depending on whether evaluation metrics target raw text references or intermediate ASR transcriptions.

\subsection{Per-Language Variance Analysis}
\label{app:radar_plots}
\begin{figure*}
    \centering
    \includegraphics[width=\linewidth]{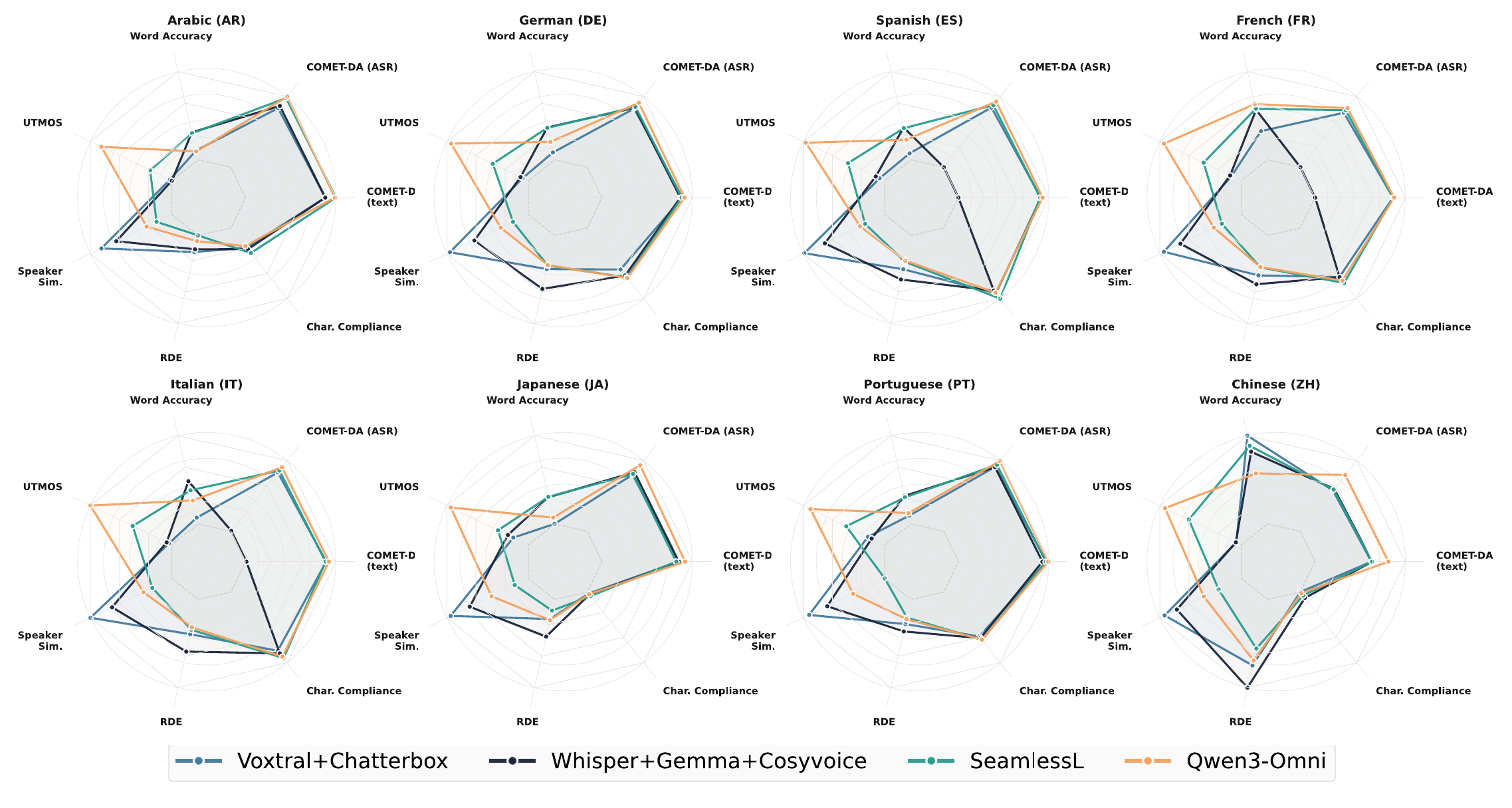}
    \caption{Per-language evaluation profiles across COMPASS dimensions on the CVSS X$\rightarrow$EN dataset. These visualizations demonstrate the structural generalization of the compact metric subset under different corpus conditions. The plots expose localized variances in speech-length compliance and translation quality metrics, confirming that the orthogonal axes effectively isolate semantic accuracy from pure acoustic naturalness without cross-dimensional metric leakage.}    \label{fig:cvss_x_en_radar_per_language}%
    \vspace{-3mm}
\end{figure*}

While the system-level averages discussed in Section~\ref{sec:results} (and more in detail in the previous section) establish broad performance trends across paradigms, they mask substantial language-specific variations. In this section, we provide a granular breakdown of metric behavior across individual target languages. These patterns are visually summarized via the per-language radar plots presented in Figs.~\ref{fig:fleurs_x_en_radar_per_language},~\ref{fig:fleurs_en_x_radar_per_language}, and~\ref{fig:cvss_x_en_radar_per_language} for FLEURS X$\rightarrow$EN, FLEURS EN$\rightarrow$X, and CVSS X$\rightarrow$EN, respectively.

\paragraph{Cascades Vulnerability to Specific Morphologies.}
The per-language distributions reveal that language distance radically affects pipeline performance compared to native end-to-end models. A prime example is observed in the CVSS $X\to\text{EN}$ direction for the three-stage cascade (\texttt{ASR+MT+TTS}). On paper, its semantic text translation quality seems competitive; however, looking at the language level exposes an extreme localized failure. For German (\texttt{de}) and Spanish (\texttt{es}), the system achieves a \texttt{COMET-DA (text)} score of 0.813 and 0.394, respectively. The high collapse in Spanish is caused by decoding loops during text generation on specific Romance languages, where the model repeatedly outputs identical phrases. Native end-to-end models like \texttt{Q3O} and \texttt{SeamL} remain completely insulated from this localized degradation, maintaining steady profiles across both language groups (e.g., maintaining $\ge 0.81$ on French and Spanish alike).

\paragraph{Acoustic / Structural Orthogonal Dynamics.}
The language breakdown further validates that voice traits and structural alignment fluctuate independently of text quality across different language families:
\begin{itemize}[noitemsep,topsep=5pt,partopsep=0pt]
    \item \textit{Speaker Consistency:} The two-stage cascade (\texttt{S2TT+TTS}) shows a flat, high-performing line for speaker identity (\texttt{Spk Sim}), holding above 0.84 across structural opposites like Arabic (\texttt{ar}: 0.849), German (\texttt{de}: 0.895), and Chinese (\texttt{zh}: 0.885). This confirms that dedicated voice-cloning TTS blocks decouple speaker preservation entirely from the complexity of target text syntax.
    \item \textit{Temporal Compression Errors:} Relative Duration Error (\texttt{RDE}) shows that temporal compression is a universal architectural feature rather than a language-specific trait. Across all ten languages on FLEURS $X\to\text{EN}$, every single system outputs consistently negative values. The magnitude of this compression varies by language family, peaking on complex character sets like Japanese (\texttt{ja}) and Korean (\texttt{ko}), where \texttt{SeamL} drops to its lowest scores of $-0.487$ and $-0.480$.
    \item \textit{Length Constraint:} Isometry matching (\texttt{Char Compl. (\%)}) reveals a noise-like distribution across language pairs. On FLEURS $\text{EN}\to X$, compliance for all systems drops below 25\% across nearly all Western European targets, yet spikes asymmetrically to over 50\% on Portuguese (\texttt{pt}) for both cascades and end-to-end setups. This highly synchronized variation across radically different systems proves that length compliance is currently driven by the natural syllable-to-character density of the target data, rather than explicit behavioral constraints inside the architectures.
\end{itemize}
The per-language breakdowns underscore the core thesis behind the COMPASS framework: evaluation setups that rely on singular, language-agnostic averages hide critical engineering trade-offs. Comprehensive speech translation profiling requires tracking these metrics down to localized language groupings.

\begin{figure*}[t]
  \includegraphics[width=2\columnwidth]{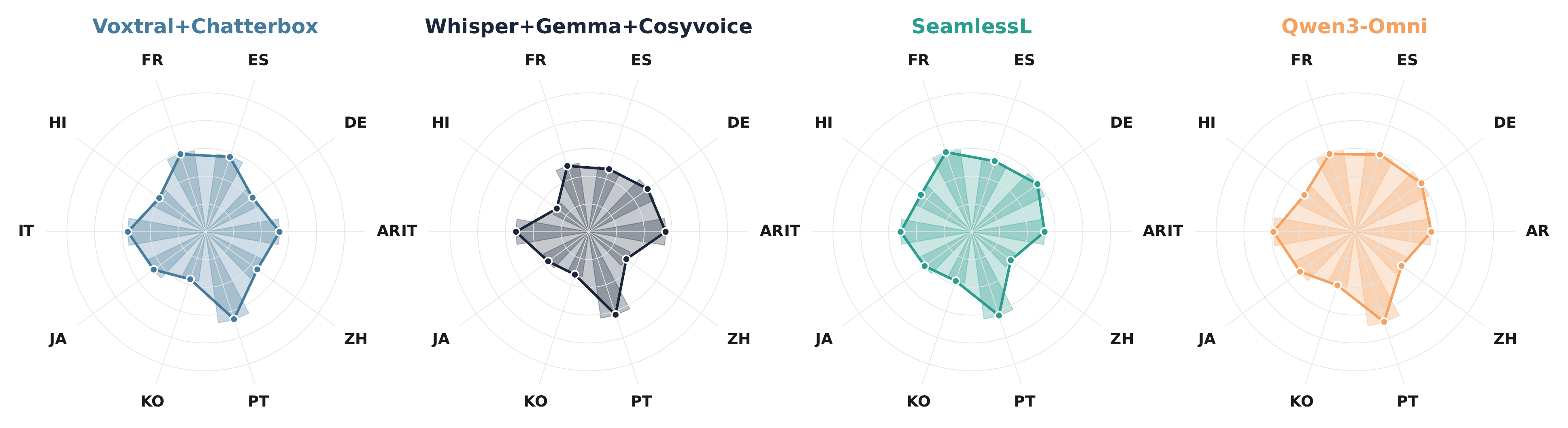}
  \caption{Language Proficiency Profile, FLEURS EN$\rightarrow$X.}
  \label{fig:petal_enx}
\end{figure*}

\begin{table}[!t]
     \centering
     \resizebox{\linewidth}{!}{%
     \begin{tabular}{rllc}
         \toprule
         \multicolumn{2}{c}{\textbf{Data}}
            & \textbf{System} 
            & \textbf{CV} \\
         \midrule
         \multirow{4}{*}{\rotatebox{90}{\footnotesize FLEURS}}
             & \multirow{4}{*}{\rotatebox{90}{\footnotesize X$\rightarrow$EN}}
         & Voxtral+Chatterbox       & 29.5\% \\
         && Whisper+Gemma+CosyVoice  & 13.5\% \\
         && SeamlessL                & 26.9\% \\
         && Qwen3-Omni               & 19.5\% \\
         \midrule
         \multirow{4}{*}{\rotatebox{90}{\footnotesize FLEURS}}
             & \multirow{4}{*}{\rotatebox{90}{\footnotesize EN$\rightarrow$X}}
         & Voxtral+Chatterbox       & 18.2\% \\
         && Whisper+Gemma+CosyVoice  & 25.5\% \\
         && SeamlessL                & 20.2\% \\
         && Qwen3-Omni               & 16.8\% \\
         \midrule
         \multirow{4}{*}{\rotatebox{90}{\footnotesize CVSS}}
             & \multirow{4}{*}{\rotatebox{90}{\footnotesize X$\rightarrow$EN}}   
         & Voxtral+Chatterbox       &  9.8\% \\
         && Whisper+Gemma+CosyVoice  & 20.8\% \\
         && SeamlessL                & 15.8\% \\
         && Qwen3-Omni               & 10.6\% \\
         \bottomrule
     \end{tabular}}
     \caption{Composite score dispersion across individual languages grouped per system and evaluation direction. A higher CV percentage reveals lower cross-lingual ranking stability.}
     \label{tab:dispersion}
\end{table}

\subsection{Extended Language Proficiency Analysis}
\label{app:proficiency}
To quantify the performance irregularity visible across distinct language targets, we compute the Coefficient of Variation (CV) of the composite system quality scores across all evaluated language branches. The metric is defined as $\text{CV} = \frac{\sigma}{\mu}$, where $\sigma$ indicates the standard deviation of the cross-lingual composite scores and $\mu$ denotes their corresponding mean value. A higher CV indicates that a system's evaluation performance depends heavily on the specific target or source language branch, reflecting weaker cross-lingual generalization capabilities. 

Table~\ref{tab:dispersion} reports the CV configurations for each system across both the FLEURS and CVSS datasets.
Three main patterns emerge from this dispersion analysis, expanding upon the visual profiles presented in the main text (Fig.~\ref{fig:petal_xen}, X$\rightarrow$EN) and here with Fig.~\ref{fig:petal_enx} for EN$\rightarrow$X. 

First, the cascaded \texttt{Whisper+Gemma+CosyVoice} pipeline achieves the lowest overall CV on FLEURS X$\to$EN (13.5\%), indicating a highly uniform and stable cross-lingual performance profile when converting speech back into English. However, its CV nearly doubles when translating in the inverse EN$\to$X direction (25.5\%), a shift driven by high score drops on complex target languages like Hindi and Korean (visualized in the expanded petal layouts of Fig.~\ref{fig:petal_enx}). Interestingly, on CVSS X$\to$EN, this same cascade exhibits a much higher variance (20.8\%) than on FLEURS, showing that its performance stability is highly corpus-dependent.

Second, the end-to-end \texttt{Qwen3-Omni} architecture exhibits a highly robust profile across both benchmarks. While its variance is slightly higher on FLEURS X$\to$EN (19.5\%) than on EN$\to$X (16.8\%), it yields an exceptionally tight and stable cross-lingual spread on CVSS X$\to$EN with a CV of just 10.6\%, matching the tight symmetrical geometry shown in the main text's petal plots.

Third, \texttt{Voxtral+Chatterbox} demonstrates an inconsistent footprint across datasets and directions. It has a very high CV on FLEURS X$\to$EN (29.5\%), which drops down to 18.2\% on EN$\to$X, yet it secures the lowest overall variance on CVSS X$\to$EN (9.8\%). This confirms that its language proficiency profile fluctuates more aggressively when handling diverse, noisier acoustic inputs (FLEURS) than when it is evaluating cleaner corpora (CVSS).

These directional and corpus-driven asymmetries further complement our findings in RQ2 by proving that X$\to$EN and EN$\to$X tracks present fundamentally distinct evaluation bottlenecks, where the specific source or target language directly shifts the performance variance of the systems.

\section{Human Evaluation: Detailed Setup}
\label{app:human_eval}
This appendix provides the full details of the human evaluation reported in Section~\ref{sec:human_eval} of the main document, including data construction, annotation protocol, and per-domain results.

\subsection{Data Construction}
\label{app:human_eval_data}

\paragraph{Dubbing (MELD-ST).}
We build the dubbing evaluation set from MELD-ST~\cite{chen2024meld}, focusing on the EN$\to$JA and JA$\to$EN directions.
We start from the official MELD-ST alignment and group utterances by dialogue.
For each dialogue, we compute the cumulative duration as the sum of its utterance durations and rank dialogues from longest to shortest.
We then compute a difficulty score per dialogue, defined as the unweighted mean of four normalized components:
(i) the proportion of utterances labeled with a hard emotion (\textit{anger}, \textit{fear}, \textit{sadness}, \textit{disgust});
(ii) the proportion of utterances with non-neutral sentiment;
(iii) the average word count per utterance; and
(iv) the average utterance duration.
Each component is min-max normalized to $[0, 1]$ before averaging.
From the top-100 most difficult dialogues, we retain those with a cumulative duration between 15 and 25 seconds and select 25 dialogues per direction ($\approx$ 133 atomic samples), for a total of 50 clips.
We manually verify that the EN and JA versions are correctly aligned at the dialogue level before annotation.

\paragraph{Podcasts (EuroParl).}
We construct the podcast evaluation set from audio published by the European Parliament Multimedia Centre.\footnote{\href{https://multimedia.europarl.europa.eu/en/audio}{\texttt{multimedia.europarl.europa.eu}}}
We focus on four language pairs in both directions: EN$\leftrightarrow$DE, EN$\leftrightarrow$ES, EN$\leftrightarrow$FR, and EN$\leftrightarrow$IT.
For each language pair, we download six random podcast episodes, manually align the source and target audio at the segment level, and extract 30-second segments. We verify that these segments do not contain multiple speakers or background music.
We select 6 segments per direction, yielding 12 segments per language pair and 48 clips in total.

\paragraph{Medical conversations (MultiMed-ST).}
We build the medical evaluation set from MultiMed-ST~\cite{le2025multimed}, focusing on the EN$\to$ZH and ZH$\to$EN directions.
We select 20 segments per direction, each between 8 and 14 seconds long, for a total of 40 clips.
Each segment is manually inspected against four inclusion criteria:
(i) the segment contains either the clinician or the patient speaking, but not both;
(ii) there is no musical background;
(iii) there is no podcast-style overlapping speech;
(iv) at least one medical term is present in the segment.
Segments that fail any of these criteria are discarded and replaced.

\paragraph{System Selection for Human Evaluation.}
To provide a representative cross-section of current speech-to-speech technology, our human evaluation covers the same four core systems analyzed throughout our automatic benchmarks. These configurations deliberately span distinct architectural approaches and quality tiers: the end-to-end SpeechLLM \texttt{Qwen3-Omni} (referred to as \texttt{Q3O}), the dedicated end-to-end \texttt{SeamlessM4T-Large-v2} (referred to as \texttt{SeamL}), and two cascading pipelines representing traditional structural boundaries, the two-stages \texttt{Voxtral+Chatterbox} (\texttt{S2TT+TTS}) and the three-stages \texttt{Whisper-Lv3+Gemma-3+Cosyvoice3} (\texttt{ASR+MT+TTS}) suite. Evaluating this specific quadrant allows us to directly cross-reference human qualitative judgments against the stable and compact metric dimensions identified by the COMPASS pipeline.
Additionally, human evaluators score the original ground-truth reference audio alongside the system outputs across randomized trials, with the sole exception of the medical domain, where reference speech tracks are unavailable.

\subsection{Annotation Protocol}
\label{app:human_eval_protocol}

\paragraph{Interface.}
Annotation is performed through a custom web-based interface that presents, for each item, the source audio and the target output.\footnote{For the medical domain, the reference text translation as well, while for dubbing they are provided with the entire video.} Annotators can replay each audio file as many times as needed before submitting their ratings. Before annotation, all participants are provided with detailed guidelines defining each evaluation axis, the corresponding 5-point Likert scale anchors, and worked examples illustrating typical low-, mid-, and high-quality outputs to ensure consistent interpretation across annotators.

\paragraph{Annotator Profile.} Each evaluation item is rated independently by three distinct annotators. All annotators are native speakers of the target non-English language and possess verified business-level English proficiency to ensure reliable source-to-target quality assessments.

\paragraph{Rating Framework.} Multi-dimensional quality criteria are scored using standard 5-point Likert scales with explicit behavioral anchors (ranging from 1 for very poor to 5 for excellent). To evaluate practical utility, downstream real-world applicability questions use a streamlined, action-oriented 3-point categorical scale (No / Maybe / Yes). All presentation orders and system assignments are fully randomized during annotation to mitigate position bias. Finally, annotators are asked to explicitly select the best and second-best overall systems for each evaluation instance.

\paragraph{Dubbing Evaluation.} 
For the entertainment domain, annotators assess alignment by scoring both fine-grained audio-visual synchronization features and contextual translation fidelity:
\begin{enumerate}[noitemsep,topsep=5pt,partopsep=0pt]
    \item[\textbf{Q1}] \textit{Lip Sync}: ``Does the speech match mouth movements?''
    \item[\textbf{Q2}] \textit{Voice \& Audio Quality}: ``Does the dubbed voice sound natural and clean (no artifacts)?''
    \item[\textbf{Q3}] \textit{Speaker Consistency}: ``Does the dubbed voice stay consistent throughout the dialogue?''
    \item[\textbf{Q4}] \textit{Prosody \& Rhythm}: ``Is the intonation, rhythm, and pacing natural?''
    \item[\textbf{Q5}] \textit{Timing \& Duration}: ``Does the speech duration fit the scene without awkward gaps or overlaps?''
    \item[\textbf{Q6}] \textit{Speaker Similarity}: ``Does the dubbed voice match the original speaker?''
    \item[\textbf{Q7}] \textit{Emotional Match}: ``Does the dubbed speech convey the same emotion and intensity as the original scene?''
    \item[\textbf{Q8}] \textit{Translation Accuracy}: ``Is the meaning accurately and completely conveyed?''
    \item[\textbf{Q9}] \textit{Overall Quality}: ``What is your overall impression of the dubbing quality of this clip?''
\end{enumerate}

\paragraph{Podcast Evaluation.} For the media broadcasting domain, the questions focus on long-form audio clarity, vocal identity preservation, and the acoustic naturalness required for clean listening experiences:
\begin{enumerate}[noitemsep,topsep=5pt,partopsep=0pt]
    \item[\textbf{Q1}] \textit{Naturalness / Listening Comfort}: ``How natural and comfortable does the dubbed audio feel to listen to?''
    \item[\textbf{Q2}] \textit{Prosody and Expressiveness}: ``Does the translated speech sound like natural spoken language with appropriate intonation, rhythm, and emotion?''
    \item[\textbf{Q3}] \textit{Speaker Consistency}: ``Does the speaker's voice remain consistent in tone, pitch, and quality throughout the segment?''
    \item[\textbf{Q4}] \textit{Content Understanding}: ``After listening to the dubbed audio, could you follow and understand the topic being discussed?''
    \item[\textbf{Q5}] \textit{Translation Accuracy}: ``Is the meaning of the original content accurately preserved in the translation?''
    \item[\textbf{Q6}] \textit{Continued Listening}: ``Would you listen to a full podcast episode (30+ minutes) translated like this?'' (\textit{yes / no})
    \item[\textbf{Q7}] \textit{Overall Quality}: ``What is your overall impression of the quality of this dubbed podcast segment?''
\end{enumerate}

\paragraph{Medical Evaluation.} For the healthcare domain, our questions focus on translation accuracy, technical medical terminology, and whether the generated speech is safe and clear enough to be used in a real doctor-patient conversation:
\begin{enumerate}[noitemsep,topsep=5pt,partopsep=0pt]
    \item[\textbf{Q1}] \textit{Voice \& Audio Quality}: ``Does the translated speech sound natural and clean (no artifacts)?''
    \item[\textbf{Q2}] \textit{Prosody and Expressiveness}: ``Does the translated speech sound like natural spoken language with appropriate intonation, rhythm, and emotion?''
    \item[\textbf{Q3}] \textit{Speaker Similarity}: ``Does the translated voice match the original speaker's voice characteristics?''
    \item[\textbf{Q4}] \textit{Translation Accuracy}: ``Is the meaning of the original content accurately preserved in the translation?''
    \item[\textbf{Q5}] \textit{Medical Terminology Accuracy}: ``Are medical terms, dosages, symptoms, and instructions correctly conveyed?''
    \item[\textbf{Q6}] \textit{Real-World Usability / Trust}: ``Would you trust and use this translation service in a real doctor-patient conversation where you are the patient?'' (\textit{no / maybe / yes})
    \item[\textbf{Q7}] \textit{Overall Quality}: ``What is your overall impression of the quality of this medical conversation segment?''
\end{enumerate}

\begin{table}[t]
    \centering
    \resizebox{\columnwidth}{!}{%
    \begin{tabular}{lrrr}
    \toprule
    \textbf{Q} 
        & \textbf{JA$\to$EN} 
        & \textbf{EN$\to$JA} 
        & \textbf{Mean} \\
    \midrule
    Q1 Lip Sync            & 0.670          & 0.692             & 0.681 \\
    Q2 Voice \& Audio      & 0.817          & 0.791             & 0.804 \\
    Q3 Speaker Consist.    & 0.811          & 0.841             & 0.826 \\
    Q4 Prosody \& Rhythm   & 0.763          & 0.767             & 0.765 \\
    Q5 Timing \& Duration  & 0.691          & 0.753             & 0.722 \\
    Q6 Speaker Similarity  & 0.849          & 0.819             & 0.834 \\
    Q7 Emotional Match     & 0.804          & 0.776             & 0.790 \\
    Q8 Translation Acc.    & 0.841          & 0.857             & 0.849 \\
    Q9 Overall Quality     & 0.885          & 0.823             & 0.854 \\
    \midrule
    Pref. $\kappa$ (avg)   & 0.812          & 0.840             & 0.826 \\
    \bottomrule
    \end{tabular}}
    \caption{Krippendorff's $\alpha$ per question, MELD-ST dubbing. The ``Mean'' columns display the average across the two translation directions.}
    \label{tab:iaa_dubbing}
\end{table}

\begin{table}[!ht]
    \centering
    \resizebox{\columnwidth}{!}{%
    \begin{tabular}{lccccc}
    \toprule
    \textbf{Question} & \textbf{DE$\to$EN} & \textbf{ES$\to$EN} & \textbf{FR$\to$EN} & \textbf{IT$\to$EN} & \textbf{Mean} \\
    \midrule
    \multicolumn{6}{c}{\textit{Source-to-English (X$\to$EN)}} \\
    \midrule
    Q1 Naturalness          & 0.93 & 0.94 & 0.93 & 0.94 & 0.94 \\
    Q2 Prosody              & 0.93 & 0.95 & 0.93 & 1.00 & 0.95 \\
    Q3 Speaker Consist.     & 0.89 & 0.88 & 1.00 & 0.93 & 0.93 \\
    Q4 Content Underst.     & 0.94 & 0.95 & 0.94 & 0.95 & 0.95 \\
    Q5 Translation Acc.     & 0.89 & 0.91 & 0.90 & 0.87 & 0.89 \\
    Q6 Speaker Simil        & 0.95 & 0.93 & 0.94 & 0.94 & 0.94 \\
    Q7 Overall Quality      & 1.00 & 1.00 & 1.00 & 1.00 & 1.00 \\
    \midrule
    Pref. $\kappa$ (avg)    & 1.00 & 1.00 & 1.00 & 0.96 & 0.99 \\
    \midrule
    \midrule
    \textbf{Question} & \textbf{EN$\to$DE} & \textbf{EN$\to$ES} & \textbf{EN$\to$FR} & \textbf{EN$\to$IT} & \textbf{Mean} \\
    \midrule
    \multicolumn{6}{c}{\textit{English-to-Target (EN$\to$X)}} \\
    \midrule
    Q1 Naturalness          & 0.90 & 0.87 & 0.97 & 0.94 & 0.92 \\
    Q2 Prosody              & 0.94 & 0.92 & 0.98 & 0.96 & 0.95 \\
    Q3 Speaker Consist.     & 0.96 & 0.93 & 0.98 & 0.94 & 0.95 \\
    Q4 Content Underst.     & 0.98 & 0.92 & 1.00 & 0.99 & 0.97 \\
    Q5 Translation Acc.     & 0.91 & 0.94 & 1.00 & 0.94 & 0.95 \\
    Q6 Speaker Simil.       & 0.95 & 0.93 & 0.98 & 0.94 & 0.95 \\
    Q7 Overall Quality      & 0.99 & 0.95 & 1.00 & 0.98 & 0.98 \\
    \midrule
    Pref. $\kappa$ (avg)    & 0.98 & 0.92 & 1.00 & 1.00 & 0.98 \\
    \bottomrule
    \end{tabular}}
    \caption{Krippendorff's $\alpha$ per evaluation question on the Multimedia Europarl data, stratified by translation direction ($X\to\text{EN}$ vs. $\text{EN}\to X$). Q6 evaluates speaker similarity using a three-point scale (no/maybe/yes). The ``Mean'' columns display the average within each respective directional block.}
    \label{tab:iaa_podcasts}
\end{table}

\begin{table}[!ht]
    \centering
    \resizebox{\columnwidth}{!}{%
    \begin{tabular}{lrrr}
    \toprule
    \textbf{Q} & \textbf{ZH$\to$EN} & \textbf{EN$\to$ZH} & \textbf{Mean} \\
    \midrule
    Q1 Voice Quality
        & 0.803         
        & 0.835    
        & 0.819 \\
    Q2 Prosody        
        & 0.782          
        & 0.744  
        & 0.763 \\
    Q3 Speaker Similarity
        & 0.905 
        & 0.859    
        & 0.882 \\
    Q4 Translation Accuracy
        & 0.846          
        & 0.880  
        & 0.863 \\
    Q5 Medical Term. Accuracy
        & 0.843          
        & 0.815             
        & 0.829 \\
    Q6 Real-World Usability
        & 0.876
        & 0.910    
        & 0.893 \\
    Q7 Overall Quality
        & 0.908
        & 0.814    
        & 0.861 \\
    \midrule
    Pref. $\kappa$ (avg)
        & 0.855          
        & 0.857             
        & 0.856 \\
    \bottomrule
    \end{tabular}}
    \caption{Krippendorff's $\alpha$ per question, MultiMed-ST medical. Q6 uses a three-point scale (no/maybe/yes).}
    \label{tab:iaa_medical}
\end{table}

\subsection{Inter-Annotator Agreement}
\label{app:human_eval_iaa}
Tables~\ref{tab:iaa_dubbing}, \ref{tab:iaa_podcasts}, and \ref{tab:iaa_medical} provide the granular breakdown of Krippendorff's $\alpha$ across all evaluation questions, language pairs, and translation directions. We additionally report the mean pairwise Cohen's $\kappa$ to represent agreement on explicit preference judgments.

\paragraph{Interpretation Thresholds.}
We interpret annotation reliability following standard content analysis paradigms~\cite{krippendorff2011agreement}. Coefficients where $\alpha \geq 0.80$ indicate highly reliable consensus; ranges where $0.67 \leq \alpha < 0.80$ denote tentatively reliable agreement; and values where $\alpha < 0.67$ represent weaker baseline alignment.

\paragraph{Domain-Specific Profiles.}
As summarized in the main text, human consensus is tightly coupled to the underlying communicative objectives of each target domain:
\begin{itemize}[noitemsep,topsep=5pt,partopsep=0pt]
    \item \textit{EuroPodcasts:} Speech evaluation within clean, narrative media conditions yields the most robust agreement profiles across the entire benchmark, with directional averages of scores spanning $\alpha = 0.89\text{-}1.00$. Annotators achieve near-perfect consensus on overall quality ($\alpha = 0.98\text{-}1.00$) and high consistency on content understanding ($\alpha = 0.95\text{-}0.97$).
    \item \textit{MultiMed-ST (Medical):} Annotations within high-stakes clinical contexts remain consistently robust, with dimensional means tightly bound between $\alpha = 0.76$ and $\alpha = 0.90$. The highest agreement is on indicators like real-world usability ($\alpha = 0.90$), while it drops at its lowest on acoustic prosody ($\alpha = 0.76$).
    \item \textit{MELD-ST (Dubbing):} Multimodal video dubbing presents a much lower baseline agreement overall ($\alpha = 0.68\text{-}0.85$). While high-level metrics like translation accuracy ($\alpha = 0.85$) and holistic overall quality ($\alpha = 0.85$) track with high stability, human consensus degrades substantially when evaluating highly granular temporal dimensions such as lip synchronization ($\alpha = 0.68$) and structural timing/duration alignment ($\alpha = 0.72$).
\end{itemize}
These contrasting baselines emphasize that human evaluation data cannot be treated as a uniform gold standard, as its reliability changes inherently with the task complexity of each domain.

\paragraph{Cross-Domain Insights.}
Three insights emerge from analyzing these agreements jointly. 
First, high-level questions like \textit{overall quality} consistently get the highest agreement scores in every single domain. This proves that humans are great at judging global quality, making these scores highly reliable anchors for testing automated metrics.
Second, prosodic elements (like rhythm, expression, and intonation) are the hardest traits for human judges to agree on, except in perfectly clean audio environments like podcasts. This difficulty makes automated prosody metrics incredibly valuable.
Finally, judges align much better on high-level communicative goals, such as whether a medical translation is safe for \textit{real-world usability}, than on fine-grained audio details like naturalness. This shows that humans find it easier to agree on whether a system successfully gets the job done rather than scoring its exact acoustic features.

\subsection{Metric-Human Correlation: Full Results}
\label{app:human_eval_corr}
We report Spearman $\rho$ between system-level COMPASS metric scores and averaged human ratings for each question and each domain.
We focus on the global summaries (aggregated over all language pairs within a domain) for readability; per-language-pair results are qualitatively consistent.
Given $n = 4$-$5$ systems per correlation, all $\rho$ values should be interpreted as indicative of metric behaviour rather than statistically confirmed (see discussion in the main paper).

\begin{table}[!ht]
    \centering
    \resizebox{\columnwidth}{!}{%
    \begin{tabular}{rlll}
    \toprule
    \textbf{Rank} & \textbf{Metric} & \textbf{Avg $|\rho|$} & \textbf{COMPASS dimension} \\
    \midrule
    1  & \texttt{COMET-DA (ASR)}              & 0.82 & Translation Quality (ASR) \\
    2  & \texttt{Semantic Score (ASR)}        & 0.79 & Translation Quality (ASR) \\
    3  & \texttt{CPS Ratio}                   & 0.77 & Isometry \\
    4  & \texttt{Chars LR}                    & 0.75 & Isometry \\
    5  & \texttt{F0 Contour Similarity}       & 0.74 & Prosody \& Emotion \\
    6  & \texttt{TER (ASR)}                   & 0.73 & Translation Quality (ASR) \\
    7  & \texttt{AutoPCP}                     & 0.71 & Prosody \& Emotion \\
    8  & \texttt{COMET-DA (Text)}             & 0.71 & Translation Quality (Text) \\
    9  & \texttt{COMET-KIWI (ASR)}            & 0.71 & Translation Quality (ASR) \\
    10 & \texttt{Speech Overlap}              & 0.71 & Isochrony \\
    \bottomrule
    \end{tabular}}
    \caption{Top-10 automatic metrics by average $|\rho|$ across \textit{all} questions and \textit{all} three domains. This is the most robust view of metric utility.}
    \label{tab:global_top10}
    \vspace{-3mm}
\end{table}
\paragraph{Methodology.}
For each domain, we average the three annotators' scores per question per system per clip, then aggregate to the system level by averaging over clips within each language pair.
We then compute Spearman $\rho$ between the resulting system-level human scores and the corresponding COMPASS metric scores.

\paragraph{Metrics that consistently fail.}
Several widely used metrics show near-zero or negative correlation with human judgment across all domains and all questions.
\texttt{UTMOS} is a near-zero predictor in podcasts, a negative predictor for emotional match in dubbing ($\rho = -0.14$), and uncorrelated in medical ($\rho \approx 0.00$ for overall quality).
\texttt{NISQA-MOS} shows similar patterns, correlating strongly \textit{negatively} with emotional match in dubbing ($\rho = -0.90$): systems with perceptually clean synthesis strip out expressive prosody.
\texttt{Speaker Similarity} is weakly correlated across all three domains.
These results suggest that MOS-based audio quality metrics and raw speaker similarity measures, while useful in TTS evaluation, are poor proxies for S2ST quality as perceived by human listeners across diverse domains.

\paragraph{The isochrony paradox in dubbing.}
Lip Syncing and Timing \& Duration are the least reliable human dimensions in dubbing ($\alpha = 0.68$ and $0.72$ across both directions, respectively), yet automatic timing metrics achieve perfect or near-perfect correlation with averaged human scores: \texttt{$\Delta$ Duration} and \texttt{CPS} reach $\rho = -1.00$ for Q5 in the EN$\to$JA direction.
This indicates that individual annotators cannot reliably judge whether speech duration fits the scene, but their \textit{average} score still tracks the ground truth ordering because timing errors produce consistent discomfort even when annotators cannot articulate why.
Automatic metrics measure the underlying signal directly and without perceptual noise, making them most valuable precisely where human judgment is most variable.

\paragraph{Direction effects in podcasts.}
For X$\to$EN podcast pairs, the top metrics by average $|\rho|$ across all questions are \texttt{ChrF++ (ASR)} and \texttt{COMET-DA (ASR)} (all 0.91).
For EN$\to$X, the same five metrics lead (all $\approx$ 0.85).
The metric set is symmetric across directions for podcasts, but the absolute correlation values are slightly higher in the X$\to$EN direction, consistent with higher annotator agreement in that direction (annotators evaluate output in their native language).

\paragraph{Direction effects in dubbing.}
The top metrics for JA$\to$EN dubbing are \texttt{$\Delta$ Duration}, \texttt{COMET-DA (ASR)}, \texttt{Tempo Ratio}, and \texttt{Speech Length Compliance}.
For EN$\to$JA, the same five metrics lead but with a higher average $|\rho|$ (0.756 vs 0.711), and \texttt{CPS Ratio} replaces \texttt{Tempo Ratio} as the second-ranked metric.
The shift reflects a difference in bottleneck: in JA$\to$EN dubbing, the primary failure mode is duration mismatch of the synthesized English; in EN$\to$JA, characters-per-second ratio captures the additional compression challenge of rendering English into Japanese mora-timed speech.

\paragraph{Direction effects in medical.}
For ZH$\to$EN, \texttt{BLASER} leads for audio quality and \texttt{COMET-DA (Text/ASR)} leads for translation quality.
For EN$\to$ZH, \texttt{COMET-DA (Text)} and \texttt{COMET-KIWI (Text)} are perfect predictors of overall quality ($\rho = 1.000$), while \texttt{BLASER} drops in importance.
The asymmetry reflects that translating \textit{into} Chinese is more constrained by text-level translation accuracy than translating \textit{from} Chinese, where audio-level rendering quality becomes the primary differentiator.

\paragraph{The medical terminology gap.}
No COMPASS metric shows reliable correlation with medical terminology accuracy or real-world trust across both directions.
This gap represents a genuine limitation of current automatic evaluation for medical S2ST and motivates future work on NER-based terminology scoring, medical language model-based semantic similarity, and trust-calibrated evaluation protocols.

\paragraph{Globally robust metrics.}
Table~\ref{tab:global_top10} reports the top-10 automatic metrics ranked by average $|\rho|$ across all questions and all three domains, providing the most aggregated view of metric utility in our benchmark. 
Translation-quality metrics dominate the ranking, with \texttt{COMET-DA (ASR)} and \texttt{Semantic Score (ASR)} leading at $|\rho| = 0.82$ and $0.79$, respectively, confirming that semantic fidelity is the single most universally predictive dimension of human judgment. 
Isometry (\texttt{CPS Ratio}, \texttt{Chars LR}) and prosody (\texttt{F0 Contour Similarity}, \texttt{AutoPCP}) metrics follow closely, indicating that length control and prosodic alignment provide complementary 
signal beyond pure translation accuracy. 
Importantly, no MOS-based or speaker-similarity metric appears in the top-10, reinforcing that acoustic-quality predictors do not generalize as cross-domain indicators of S2ST quality. 
While these global rankings highlight broadly useful metrics, the per-domain analyses above show that the \textit{optimal} subset still varies substantially by application, motivating the domain-specific recommendations that follow.

\subsection{Domain-Specific Metric Recommendations}
\label{app:human_eval_recommendations}
Based on the correlation analysis, we derive the following domain-specific metric recommendations.

For \textit{dubbing}, the core set should include \texttt{COMET-DA (ASR)}, \texttt{AutoPCP}, \texttt{CPS Ratio}, \texttt{$\Delta$ Duration},\footnote{Interestingly, we found the median being more informative than the mean.} and \texttt{Speech Overlap}.
Lip sync dimensions where human agreement collapses should be evaluated with automatic metrics (\texttt{Viseme DTW}, \texttt{Co-Occurrence Score}).

For \textit{podcasts}, the core evaluation set should include \texttt{COMET-DA (ASR)}, \texttt{Semantic Score (ASR)}, \texttt{Chars Length Compliance}, and \texttt{AutoPCP}.
MOS-based naturalness metrics (\texttt{UTMOS}, \texttt{NISQA-MOS}) should \textit{not} be used as primary metrics for this domain.

For \textit{medical} conversations, the core set should include \texttt{COMET-DA (Text)}/\texttt{COMET-KIWI (Text)}, \texttt{BLASER}, and \texttt{Semantic Score (ASR)}.
The trust and terminology dimensions currently lack reliable automatic proxies and represent an open evaluation challenge.

These recommendations are grounded in correlation with human judgment rather than in theoretical arguments alone, and they differ substantially across domains, directly supporting the central claim of this paper: there is no universal metric set for S2ST evaluation.

\begin{table}[!ht]
    \centering
    \resizebox{\columnwidth}{!}{%
    \begin{tabular}{lrrrr}
    \toprule
    \textbf{System} & \textbf{Direction} & \textbf{First rate}
    & \textbf{Second rate} & \textbf{Avg Q9} \\
    \midrule
    Official dub          & JA$\to$EN & 1.00 & 0.00 & \textbf{4.99} \\
    S2TT+TTS              & JA$\to$EN & 0.00 & 0.60 & \underline{3.71} \\
    ASR+MT+TTS            & JA$\to$EN & 0.00 & 0.30 & 3.57 \\
    Q3O                   & JA$\to$EN & 0.00 & 0.10 & 2.59 \\
    SeamL                 & JA$\to$EN & 0.00 & 0.00 & 1.57 \\
    \midrule
    Official dub          & EN$\to$JA & 1.00 & 0.00 & \textbf{4.95} \\
    S2TT+TTS              & EN$\to$JA & 0.00 & 0.76 & \underline{2.64} \\
    ASR+MT+TTS            & EN$\to$JA & 0.00 & 0.24 & 2.01 \\
    Q3O                   & EN$\to$JA & 0.00 & 0.00 & 2.13 \\
    SeamL                 & EN$\to$JA & 0.00 & 0.00 & 1.27 \\
    \midrule
    Official dub          & Both & 1.00 & 0.00 & \textbf{4.97} \\
    S2TT+TTS              & Both & 0.00 & 0.46 & \underline{3.17} \\
    ASR+MT+TTS            & Both & 0.00 & 0.30 & 2.85 \\
    Q3O                   & Both & 0.00 & 0.14 & 2.30 \\
    SeamL                 & Both & 0.00 & 0.00 & 1.42 \\
    \bottomrule
    \end{tabular}}
    \caption{System rankings by human preference, dubbing. First rate = fraction of clips where this system was preferred. Q9 is the question regarding the overall dubbing quality. Best results in \textbf{bold}, second-best \underline{underlined}.}   
    \label{tab:rankings_dubbing}
\end{table}

\begin{table}[!ht]
    \centering
    \resizebox{\columnwidth}{!}{%
    \begin{tabular}{lrrrr}
    \toprule
    \textbf{System} & \textbf{Direction} & \textbf{First rate} 
    & \textbf{Second rate} & \textbf{Avg Q7} \\
    \midrule
    Ground truth          & X$\to$EN & 0.50 & 0.28 & \underline{4.72} \\
    S2TT+TTS              & X$\to$EN & 0.50 & 0.28 & \textbf{4.89} \\
    ASR+MT+TTS            & X$\to$EN & 0.00 & 0.33 & 3.88 \\
    Q3O                   & X$\to$EN & 0.00 & 0.11 & 3.72 \\
    SeamL                 & X$\to$EN & 0.00 & 0.00 & 2.33 \\
    \midrule
    Ground truth          & EN$\to$X & 0.89 & 0.06 & \textbf{4.89} \\
    S2TT+TTS              & EN$\to$X & 0.06 & 0.28 & 3.49 \\
    ASR+MT+TTS            & EN$\to$X & 0.05 & 0.39 & \underline{3.88} \\
    Q3O                   & EN$\to$X & 0.00 & 0.21 & 2.85 \\
    SeamL                 & EN$\to$X & 0.00 & 0.06 & 2.85 \\
    \midrule
    Ground truth          & Both     & 0.69 & 0.17 & \textbf{4.81} \\
    S2TT+TTS              & Both     & 0.28 & 0.28 & \underline{4.19} \\
    ASR+MT+TTS            & Both     & 0.03 & 0.36 & 3.88 \\
    Q3O                   & Both     & 0.00 & 0.16 & 3.29 \\
    SeamL                 & Both     & 0.00 & 0.03 & 2.59 \\
    \bottomrule
    \end{tabular}}
    \caption{System rankings by human preference, podcasts. Q7 is the question regarding the overall quality. Best results in \textbf{bold}, second-best \underline{underlined}.}
    \label{tab:rankings_podcasts}
\end{table}

\begin{table}[!ht]
    \centering
    \resizebox{\columnwidth}{!}{%
    \begin{tabular}{lrrrr}
    \toprule
    \textbf{System} & \textbf{Direction} & \textbf{First rate} & \textbf{Avg Q7} \\
    \midrule
    S2TT+TTS                & ZH$\to$EN & 0.20 & 3.42 \\
    ASR+MT+TTS              & ZH$\to$EN & 0.60 & \textbf{4.24} \\
    Q3O                     & ZH$\to$EN & 0.20 & \underline{3.53} \\
    SeamL                   & ZH$\to$EN & 0.00 & 2.22 \\
    \midrule
    S2TT+TTS                & EN$\to$ZH & 0.07 & 3.20 \\
    ASR+MT+TTS              & EN$\to$ZH & 0.80 & \textbf{4.29} \\
    Q3O                     & EN$\to$ZH & 0.13 & \underline{3.76} \\
    SeamL                   & EN$\to$ZH & 0.00 & 1.62 \\
    \midrule
    S2TT+TTS                & Both      & 0.13 & 3.31 \\
    ASR+MT+TTS              & Both      & 0.70 & \textbf{4.27} \\
    Q3O                     & Both      & 0.17 & \underline{3.64} \\
    SeamL                   & Both      & 0.00 & 1.92 \\
    \bottomrule
    \end{tabular}}
    \caption{System rankings by human preference, medical. Q7 is the question regarding the overall quality. No reference audio is available for this domain, thus we did not ask for the second-best candidate system. Best results in \textbf{bold}, second-best \underline{underlined}.}
    \label{tab:rankings_medical}
\end{table}

\subsection{System Rankings by Human Preference}
\label{app:human_eval_rankings}
Tables \ref{tab:rankings_dubbing}, \ref{tab:rankings_podcasts}, and \ref{tab:rankings_medical} report system rankings by first-preference rate, second-preference rate (where reference tracking was available), and average overall quality score per domain and direction.

\paragraph{Key Observations.}
Three main cross-domain patterns stand out from the human preference tables:
First, \texttt{SeamL} receives zero first-preference votes across every single domain and translation direction. This represents the only completely consistent last-place result in our human evaluation, directly mirroring its last-place automatic scores throughout the benchmark.
Second, the optimal system depends heavily on the specific domain. 
While the end-to-end \texttt{S2TT+TTS} framework is highly competitive with human production in clean media environments like podcasts, the three-stage cascade (\texttt{ASR+MT+TTS}) heavily dominates high-stakes settings like medical dialogues. This confirms that no single architecture is best across all deployment contexts.
Third, human judges confirm the directional asymmetry observed in our automatic benchmarks. Translating from English to a target language ($\text{EN}\to X$) is consistently more challenging for models than translating into English ($X\to\text{EN}$). This difficulty is reflected in the lower absolute quality scores and the wider performance gaps between automated systems and human-produced references.

\end{document}